\documentclass[twoside,11pt]{article}

\usepackage{blindtext}

\usepackage{xcolor}
\usepackage{amsmath} 
\usepackage{amsfonts}
\usepackage{subcaption}
\usepackage{graphicx}
\usepackage{soul}
\usepackage{booktabs}
\usepackage{wrapfig}
\usepackage{natbib}

\newcommand{\R}{\mathbb{R}}

\newcommand{\veps}{\varepsilon}
\DeclareMathOperator{\Ima}{Im}

\usepackage{lastpage}

\begin{document}

\title{Topology of Out-of-Distribution Examples in Deep Neural Networks}

\author{Esha Datta \textit{edatta@sandia.gov} \\ 
        Johanna Hennig \textit{jmhenni@sandia.gov} \\
        Eva Domschot \textit{ecdomsc@sandia.gov}\\
        Connor Mattes \textit{clmatte@sandia.gov} \\
        Michael R. Smith \textit{msmith4@sandia.gov}\\
       Sandia National Laboratories Albuquerque, NM 87185, USA
       }


\maketitle

\begin{abstract}

As deep neural networks (DNNs) become increasingly common, concerns about their robustness do as well. A longstanding problem for deployed DNNs is their behavior in the face of unfamiliar inputs; specifically, these models tend to be overconfident and incorrect when encountering out-of-distribution (OOD) examples. In this work, we present a topological approach to characterizing OOD examples using latent layer embeddings from DNNs. Our goal is to identify topological features, referred to as landmarks, that indicate OOD examples. We conduct extensive experiments on benchmark datasets and a realistic DNN model, revealing a key insight for OOD detection. Well-trained DNNs have been shown to induce a topological simplification on training data for simple models and datasets; we show that this property holds for realistic, large-scale test and training data, but does not hold for OOD examples. More specifically, we find that the average lifetime (or persistence) of OOD examples is statistically longer than that of training or test examples. This indicates that DNNs struggle to induce topological simplification on unfamiliar inputs. Our empirical results provide novel evidence of topological simplification in realistic DNNs and lay the groundwork for topologically-informed OOD detection strategies.
\end{abstract}


\section{Introduction}
In the past decade, deep neural networks (DNNs) have become ubiquitous. Deep learning techniques play a crucial role in safety-critical tasks like malware detection \citep{abdelsalam2018malware}, medical image analysis \citep{medml}, and autonomous driving \citep{autodrive}. 

DNNs are generally trained under a closed-world assumption \citep{krizhevsky2012imagenet,he2015delving}; that is, it is assumed the model will only encounter data drawn from the same distribution as its training data. This assumption rarely holds for deployed models. An autonomous vehicle might encounter an unforeseen hazard in the road. A malware detector might encounter a novel virus. When confronted with these unfamiliar inputs, known in the literature as {out-of-distribution} (OOD) examples, DNNs tend to be highly confident and incorrect \citep{goodfellow2014explaining, amodei2016concrete}. Consequently, the detection of OOD examples has become an active area of research in machine learning~\citep{hendrycks2016baseline,liang2017enhancing,liu2020energy}. 

Topological data analysis (TDA) is a growing field that leverages tools from algebraic topology to examine shape features in high-dimensional data \citep{rabadan-book}. Its theoretic foundations enable TDA to excel in settings where standard analytic tools may falter. Topological techniques are robust to local variation, can operate at multiple scales simultaneously, and has guaranteed stability bounds in the presence of noise \citep{skaf2022topological}. TDA has found success in a variety of fields, including bioinformatics \citep{rabadan-book}, dynamical systems \citep{topaz2015topological}, and more recently, machine learning \citep{zhang2024comprehensive, hensel2021survey}.

\subsection{Our contributions}

This work contributes to the growing body of topological studies on DNN structure. We extend the persistent homology approach described by \citet{naitzat} to investigate the effect of OOD examples on deep learning models. Our empirical study identifies the topological landmarks (or indications) of these examples within the latent-layer space of ResNet18 models, providing new insights to the robustness and vulnerability of DNNs in the face of unfamiliar inputs. While our methods could be viewed as an extension of \citet{naitzat}, there are key departures from the original approach that enable our approach: \begin{enumerate}
\item In a novel application, we leverage the topological properties of latent layer embeddings to characterize OOD examples. We validate our results across multiple benchmark datasets; 
\item We demonstrate that the empirical result of \citet{naitzat} extends to test datasets, multi-class classification problems, and more realistic datasets and architectures; and 
\item We demonstrate the computational feasibility of a TDA pipeline for analyzing realistic DNNs at scale.
\end{enumerate}

The rest of the paper is structured as follows. Section \ref{sec:lit} discusses recent studies at the intersection of TDA and machine learning.  Section \ref{sec:int} discusses, in more detail, the intuition of our approach and our central hypotheses. In Section \ref{sec:tdaback}, we provide a brief overview of relevant topics in TDA, including details on the efficient computation of persistent homology. Section \ref{sec:methods} describes our datasets, model architectures, and experimental methodology. Section \ref{sec:results} presents selected results, with a more exhaustive treatment in Appendix \ref{sec:app}. Finally, we discuss the impact of our findings and potential avenues for future work in Section \ref{sec:disc}. 

\section{Existing Works and Opportunities for Novel Insight}\label{sec:lit}
This section highlights studies at the intersection of TDA and ML that have informed our investigation. Readers seeking a comprehensive overview of the literature are directed to \citet{ballester-review-2023} and \citet{zia-review-2024}. 

The flexibility of algebraic topology makes it well-suited for problems in ML. Algebraic topology translates the challenge of comparing shapes into a simpler one: comparing numbers. By characterizing complex structures numerically, topology enables the differentiation of high-dimensional spaces. The notion of shape equivalence (via homological invariance) gives topological approaches robustness to noise as well \citep{skaf2022topological}. These characteristics have made computational topology successful in settings where traditional approaches struggle \citep{skaf2022topological}. 

Our study employs persistent homology (PH), which is perhaps the most well-known of the TDA methods. PH can analyze data across multiple scales simultaneously and identify meaningful global signals without obfuscating local structure \citep{skaf2022topological,lum2013extracting} Moreover, a well-developed theory of stability quantifies the extent to which PH is affected by noise \citep{pds-are-stable}. Perhaps most importantly, PH has been used to study DNNs because the topological invariants associated with neural networks appear to encode relevant quantities including validation loss \citep{rieck-early-stopping}, shallow-to-deep layer connections \citep{zheng-trojan-2021}, and (we find), the relationship of inference data with respect to the training data. We discuss the existing approaches to PH for DNN analysis below. A brief technical background on PH computation is presented in Section \ref{sec:tdaback}. 

Several authors study DNNs by interpreting the model architecture as a graph and applying TDA to this construction. In an early work, \citet{rieck-early-stopping} interpret a multi-layer perceptron (MLP) model as a multipartite graph with nodes corresponding to neurons and edges a function of training weights. The study then introduces the notion of neural persistence, which characterizes the complexity of an MLP. Another approach that melds topology and graph theory, introduced by \citet{geb1}, uses the activation graph. This is a weighted, undirected multipartite graph that can be studied as a one-dimensional simplicial complex independent of the input data. Recently, \citet{lacombe-ood-2021} show how the topology of an activation graph can be used for OOD detection and uncertainty quantification. These approaches rely on modeling DNNs as sequences of bipartite graphs which, by construction, cannot capture interactions between non-adjacent layers. Therefore, the typical graphical approach cannot be used for models containing skip connections, an increasingly common feature in contemporary DNNs.  

A contrasting but complimentary line of study examines the topology of the latent layer of neural networks. In a study that is influential to our own, \citet{naitzat} explore changes in persistent homology of training data as it passes through the layers of a neural network. A key result of this work is that a well-trained neural network induces a topological simplification upon its training data. That is, if we consider the training dataset to be representative of some complex manifold $M$, we can characterize this manifold (and the changes induced on it by the neural network) via persistent homology. This approach has been explored more recently by \citet{wheeler2021activation}, \citet{akai2021experimental}, and \citet{ferra2023importance}, among others, to study the training process and performance of DNNs. Notably, \citet{akai2021experimental} examine the penultimate fully-connected layer of a network to define a relationship between topological features and performance of DNNs. 

To the best of our knowledge, our approach towards characterizing the effects of OOD examples in a DNN by studying the topology of latent layer embeddings has not been done before. The key conclusion of our work, that DNNs cannot simplify the topology of OOD examples in the manner of training data, is also novel. Further, the computational demands of PH have historically restricted TDA studies of latent layer embeddings. Even recent works focus on fully-connected CNNs and MLPs, due to their relative computational feasibility \citep{zhang2024comprehensive}. Our work addresses a gap in the literature by applying TDA to complex, wider models and characterizing the topology of OOD examples at scale. 

\section{Intuition and Motivation}\label{sec:int}

Our work extends the key observation of \citet{naitzat} that well-trained neural networks induce a simplification of the topology of the training data. We formalize this intuition as follows, our notation matching that of \citet{naitzat}. Note that this subsection assumes some familiarity with persistent homology. Readers seeking additional background on the methods are referred to Section \ref{sec:tdaback}.

Suppose that $f: \R^p \rightarrow \R^q$ is a classification function computed by a well-trained network. Further suppose that $X$ is the set of in-distribution (ID) datapoints relative to classifier $f$. That is to say, the points in $X$ are drawn from the same manifold as the training dataset. Let the set $X_m \subseteq X$ represent the ID datapoints from class $m$. Per \citet{naitzat}, as $X_m$ passes through the layers of the neural network, its homology should become increasingly trivial. That is, we expect $f(X_m)$ to consist of only one connected component with no higher-dimensional homology features. Put formally in terms of Betti numbers $\beta_k$ for $k \in \mathbb{N}$, we expect that $\beta_0(f(X_m)) = 1$ and $\beta_k(f(X_m)) = 0$ for all $k > 0$ . This observation is shown to hold true for simulation data and low-dimensional representations of MNIST in \citet{naitzat}. 

We offer two conjectures that follow from this observation. 

First, the ``trivialization'' behavior observed in \citet{naitzat} is specific to a single class, but it could generalize to multiple classes as follows. Consider the multi-class problem with classes $0, 1, \ldots, n$, the well-trained classification function $f$, and the set of in-distribution datapoints $X$. We expect each class-specific subset $X_i \subseteq X$ to be trivialized by $f$, i.e. $\beta_0 (f(X_i)) = 1$ for all classes $i$. We conjecture that the entire set $X$ should be trivialized, with each subset forming a separate connected component, resulting in $\beta_0(f(X)) = n$ connected components. This would enable $f$ to correctly classify the training data, as each class is mapped to a distinct subset of the image $\Ima(f).$ To our knowledge, this has not yet been verified empirically.

Second, the contrapositive of this observation implies that if $Y \subseteq \R^p$ is a subset of points such that $f(Y)$ has \textit{nontrivial} topology, then $Y$ consists of out-of-distribution (OOD) points. That is, the points in $Y$ are \textit{not} drawn from the same manifold as the training data.

\section{TDA Background}\label{sec:tdaback}

Algebraic topology enables the quantitative study of shape features of objects in topological spaces. Its approach is unique in its flexibility; it is concerned with the relative position of points within a topological space, rather than the absolute positions (e.g., the coordinates) of points. Algebraic topology provides tools by which we may assign algebraic invariants (e.g., numbers) to objects of interest. We can then simply compare these objects by comparing the algebraic invariants associated with them, rather than trying to compare shapes. 

Here, we introduce a simplified background on algebraic topology, the Vietoris-Rips complex, and persistent homology which we leverage in our work. For a more in-depth introduction to the subject matter, we direct readers to \citet{edelsbrunner2022computational}. 

We focus on a type of topological space known as a metric space, which is a set of points with a notion of distance (called a metric) defined for each pair of points. A metric is non-negative, symmetric, and satisfies the triangle inequality. The Euclidean space of $n$ dimensions $\R^n$ equipped with the standard distance function is an example of a metric space. In our analysis, we study neural network embeddings as high-dimensional Euclidean spaces. 

We impose structure upon the metric space via a simplicial complex. In essence, we use the metric to divide the metric space into ``simple pieces'', known as simplices, and analyze the structure that arise. Those of the first few dimensions should be familiar: a $0$-simplex is a vertex, a $1$-simplex is an edge, a $2$-simplex is a triangle, and a $3$-simplex is a tetrahedron. In general, a $k$-simplex is the convex hull of $k+1$ affinely independent points. We can think of $k$-simplices as the building blocks comprising a simplicial complex. 

The Vietoris-Rips construction can then be considered the blueprint of our simplicial complex. That is to say, the Vietoris-Rips complex informs the process by which simplices are formed and connected to yield higher-order structure. It is not the only method for producing simplicial complexes (other notable methods include the Cech \citep{ghrist-barcodes} and witness complexes \citep{witness-complex}), but it is often preferred for its computational properties. For a given distance $\veps > 0$, the Vietoris-Rips complex $S_\veps$ is formed as follows: form each collection of $k+1$ data points into a $k$-simplex if the pairwise distances between the points is less that $\veps.$ For instance, \begin{enumerate}
    \item each individual data point forms its own $0$-simplex,
    \item each pair of points within $\veps$ of each other form a $1$-simplex (an edge),
    \item each triplet of points within $\veps$ of each other form a $2$-simplex (a triangle), and 
    \item a quartet of points within $\veps$ of each other pairwise form a $3$-simplex (a tetrahedron).
\end{enumerate}

\subsection{Homology}

For the sake of parsimony, we do not discuss the underlying algebraic structure that enables homology computation. Our focus is instead on building intuition around homology and Betti numbers. We refer readers interested in the theory of homology to \citet{munkres2018elements}. 

Homology enables the quantification of shape features in a simplicial complex. We can think, informally, of homology as identifying the $k$-dimensional ``holes'' within a complex. These holes need not be circular in a geometric sense. Rather, a hole in a complex $S_\veps$ is a combination of simplices that could be a boundary of a simplex, but is not. We can characterize these holes for low dimensions in familiar terms. 

We refer to $H_k$ as the $k$-th homology group. The $k$-th Betti number $\beta_k$ is the dimension of $H_k$ and, in terms of topological features, counts the number of $k$-dimensional holes. For concreteness, $\beta_0$ is the number of connected components, $\beta_1$ is the number of circles or loops, $\beta_2$ is the number of trapped volumes, and so on. A simplicial complex can be described by the infinite sequence of Betti numbers $(\beta_0, \beta_1, \beta_2, \ldots)$. The most important feature of Betti numbers (for our purposes) is that they are a topological invariant. That is, topologically equivalent spaces have the same Betti number, thereby enabling a coarse measurement of similarity across complex spaces. 

\subsection{Persistent homology}

We have provided the Vietoris-Rips construction of a simplicial complex $S_\veps$ for a fixed distance $\veps > 0$. It should be apparent that both the complex and homology are dependent of $\veps$, which begs the question: what is the optimal choice of $\veps$ for a point cloud? In most cases, we may be unaware of (or incapable of determining) the answer. The notion of \textit{persistence} allows us to sidestep the matter altogether by instead evaluating a range of $\veps$ values. 

There is a natural inclusion of Vietoris-Rips complexes that occur as $\veps$ grows. That is to say, for an increasing sequence $\veps_0 \leq \veps_1 \leq \cdots \leq \veps_m$, there is an associated sequence of simplicial complexes $S_{\veps_0} \subseteq S_{\veps_1} \subseteq \cdots \subseteq S_{\veps_m},$ called a filtration. By tracking homological features that persist over a large range of $\veps$ values, we can identify signals of the underlying topology of our point cloud. While $\veps$ is a distance value, chosen using the metric on our point cloud, we often refer to the length of persistence as a ``lifetime.'' 

\begin{figure}[h!]
    \centering
        \includegraphics[width=0.3\linewidth]{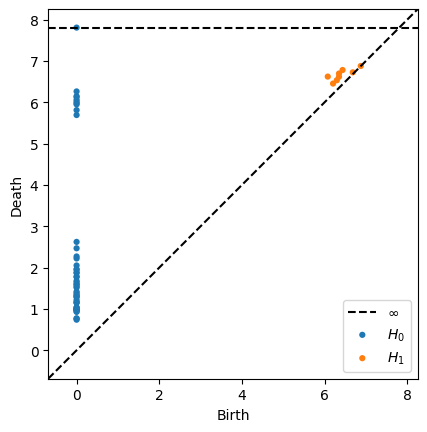}
        \caption{An example persistence diagram with birth times on the $x$-axis and death times on the $y$ axis. Points closer to the diagonal are considered noise, as their lifetime (the difference between birth and death times) is close to 0.}
        \label{fig:pds}
\end{figure}

We use persistence diagrams (Figure \ref{fig:pds}) to visualize the persistent homology of a simplicial complex in the Cartesian plane.  Each point in the diagram corresponds to a homological feature. The $x$-axis shows the start (or \textit{birth}) time of the feature, while the $y$-axis shows the end (or \textit{death}) time. Points closer to the diagonal line have shorter lifetimes and may be considered noise. 

\subsection{Computation, inference, and persistent homology}

Persistent homology (and, indeed, all of topological data analysis) asserts that data has shape and that understanding this shape can shed light on the underlying phenomena that generate this data. Research in fields ranging from genomics \citep{rabadan-book} to machine learning \citep{ballester-review-2023} has provided evidence in support of these claims. The success of persistent homology has been due, in part, to its versatility and the robust theoretic framework that has been developed to support its applications. 

The stability theorems for persistent homology provide crucial guarantees for the soundness of topological analyses. In particular, there is a known relationship between perturbations in the input dataset under the Gromov-Hausdorff distance and the perturbation of the output \citep{chazal2014persistence, pds-are-stable}. As a result, the extent to which persistent homology is noise-resistant is understood theoretically. It has also been validated empirically in computer vision, clustering, and biological applications \citep{zia-review-2024}. The stability theorems offer some assurance of the extent to which topological inference recovers the ground truth from noisy data. 

The bootstrap approach for topological data analysis has been one avenue for estimating homology over large datasets. The bootstrap is a non-parametric technique for statistical inference \citep{bootstrap}. It is used to estimate population parameters or create confidence intervals. The bootstrap has two parameters: a sample size and a number of iterations. The iterations determine how many times the dataset $X$ is resampled, while the sample size determines the number of observations selected, at random with replacement, from $X$. Typically, a statistical quantity of interest is recorded for each resample, ultimately yielding an estimate for the true value of the quantity. We refer to \citet{bootstrap} for an overview of bootstrap and its variations. \citet{rabadan-book} present a simplified explanation of approach developed by \citet{fasy2013statistical, chazal2013bootstrap} demonstrates that the empirical estimate of persistent homology from a bootstrap subsample converges (asymptotically) to the the ground truth in the $L_\infty$ norm under certain regularity assumptions.

The efficient computation of persistent homology has long been a roadblock to its widespread application. In part, this is because of the exponential rate of growth of simplicial complexes. For instance, the Vietoris-Rips complex on a finite metric space $X$ has $2^{|X|}$ simplices \citep{rabadan-book}. There are some workarounds: restricting investigation to lower-dimensional homologies, taking subsamples of tractable size, or restricting the maximum distance $\veps$ at which connections are made in a complex. Even so, the size of machine learning datasets can be a computational bottleneck, as in the work by \citet{naitzat} on the persistent homology of neural network embeddings. The authors reported that the time required to compute a simple Betti number from a point cloud $X$ drawn from MNIST ranged from a tens of seconds (for $X \subseteq \R^5$) to 30 minutes (for $X \subseteq \R^{50}$).

Recent advances in the computation of Vietoris Rips complexes have enabled investigations on much larger datasets. The Ripser algorithm \citep{ripser-paper, ripser} made substantial improvements in compute time and memory usage through its exploitation of apparent pairs. This software was further improved by \citet{rpp} with a GPU-acceleration of Ripser, which achieved an up to 30 times speedup over Ripser and reduced memory usage. While homology computation (particularly for $H_2$ and higher) remain challenging, these new software have opened the door to the application of persistent homology to problems in machine learning. 

\section{Methodology}\label{sec:methods}

We consider the effect of OOD inputs in a multi-class image classification setting. Let $M$ be the manifold from which images of interest are drawn. We assume that there exists some ideal classifier with near-zero prediction error and that the training set $T$ is sufficient for characterizing the topology of $M$. We are given a classifier $f: \R^p \rightarrow \R^q$ trained on $T$. To evaluate this classifier, we draw examples from a test set $T' \subseteq M$ and measure the accuracy of class assignment. 

Consider an example $y \in Y$ where $y \not\in M$. We refer to such examples as out-of-distribution (OOD) relative to the trained classifier $f$. Our experiments are intended to reveal the differences in topology for $f(T), f(T'),$ and $f(Y)$. 

Specifically, our experiments examine the homology of the penultimate layer of the classifier, which we refer to as the embedding layer. We measure the $H_0$ and $H_1$ homologies, compute summaries of these results, and compare them statistically. 

\subsection{Architecture and Datasets}
Our experiments are performed across four benchmark datasets from the machine learning literature: CIFAR-10, CIFAR-100, MNIST, and EMNIST. We use a ResNet18 architecture as a classifier and trained two models for image classification tasks of varying complexity: one for handwritten digit classification trained on MNIST and another for image classification trained on CIFAR-10. 

For the ResNet model trained on MNIST, we drew samples from EMNIST to serve as OOD examples. For the other, we drew from CIFAR-100 for OOD examples. 

\subsection{Persistent homology computations}

Our experiments are performed on the ResNet 18 embedding layer. As such, we analyze the persistent homology of data in $\R^{512}$ and summarize the results using: \begin{enumerate}
     \item average lifetime, which provides an estimate of the average persistence of features in the sample; 
    \item maximum lifetime, which captures any ``outliers'' or particularly persistent features in a sample; 
    \item average birth time, which indicates the $\veps$ value at which  features arise; and 
    \item average death time, which indicates the $\veps$ value at which features disappear or are coalesced into other features.  
\end{enumerate}  

All computations are performed using the Python library Ripser \citep{ripser}. By construction, the average birth time for all $H_0$ features is $\veps = 0$. This means that the average lifetime and average death time for $H_0$ are equal. This need not be the case for higher order features. Similar measures are used in the literature \citep{zheng-trojan-2021} to compare distributions of topological features in DNNs.

\subsection{Overview of Experiments}

We describe the full details of our experiments here, with specifics on bootstrap parameters, homology computation, and summary generation. Key considerations in our experimental design are: \begin{itemize}
    \item limiting the computational cost of repeated persistent homology computation, and
    \item producing a sufficiently large number of subsamples for a representative distribution of topological features. 
\end{itemize} 

We perform a descriptive investigation to estimate and compare topological summary statistics for each dataset. 

For the ID data, we separately compare the training and test data, although we do not expect those datasets to differ significantly for a well-trained model. We then compare against the corresponding OOD dataset. The procedure is as follows: for $M$ iterations, \begin{enumerate}
    \item Take a random sample, with replacement, of size $n$ from the dataset of size $N$. 
    \item Compute its persistent homology up $H_1$. 
    \item Compute the topological summary statistics for $H_0$ and $H_1$. 
\end{enumerate} This yields a distribution of each topological summary statistic for ID train, ID test, and OOD data. We can then compute a 95$\%$ confidence interval to compare the populations. For this task, we used bootstrap samples of size $25, 50, 100,$ and $150$ over $50,000$ iterations. The number of iterations was informed by computational constraints.

\section{Results}\label{sec:results}

We present the results from our experiments to analyze the topological differences between ID and OOD examples.

In addition to the ID and OOD comparisons, we compare the bootstrap results of the testing and training splits of our data. 

For simplicity, we refer to the training data (and results born from it) as Train, the testing data as Test, and the OOD as such across both MNIST and CIFAR. For each case, we vary the bootstrap sample size, increasing as much as is computationally feasible. The figures for all tested sample sizes can be found in Appendix \ref{sec:app}. 

\subsection{MNIST}

Figure \ref{fig:h0_mnist} compares topological features arising in MNIST and EMNIST. The features are measured by taking a bootstrap of size 150 from the corresponding datasets and computing the persistent homology on the subsample. The black and blue distributions represents the training and test splits, while the red represents EMNIST. Samples are taken and processed over 50000 bootstrap iterations for each distribution. 

\begin{figure}[h!]
    \centering
    \begin{subfigure}[b]{0.45\textwidth}
        \centering
        \includegraphics[width=\linewidth]{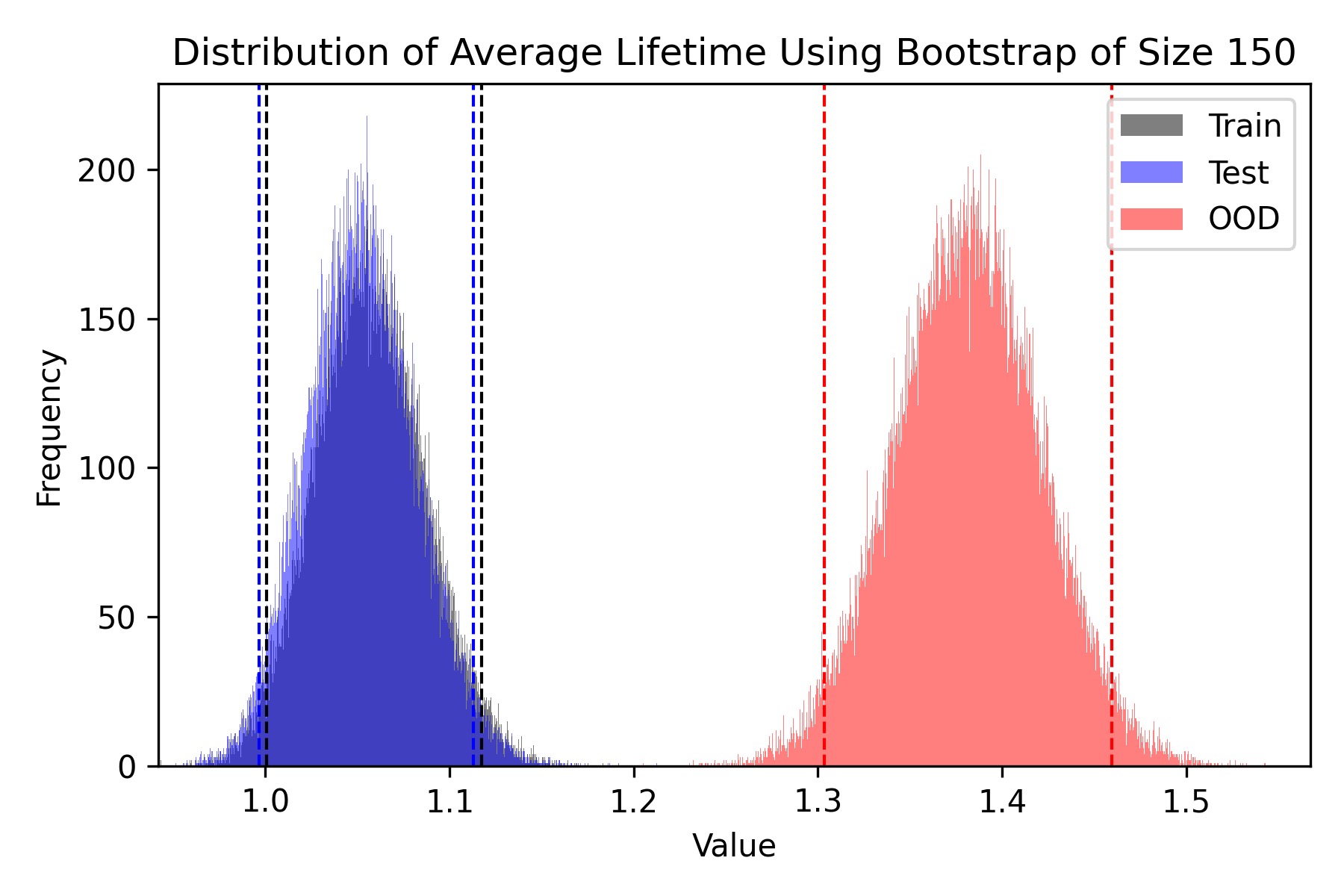}
    \end{subfigure}
    \hfill
    \begin{subfigure}[b]{0.45\textwidth}
        \centering
        \includegraphics[width=\linewidth]{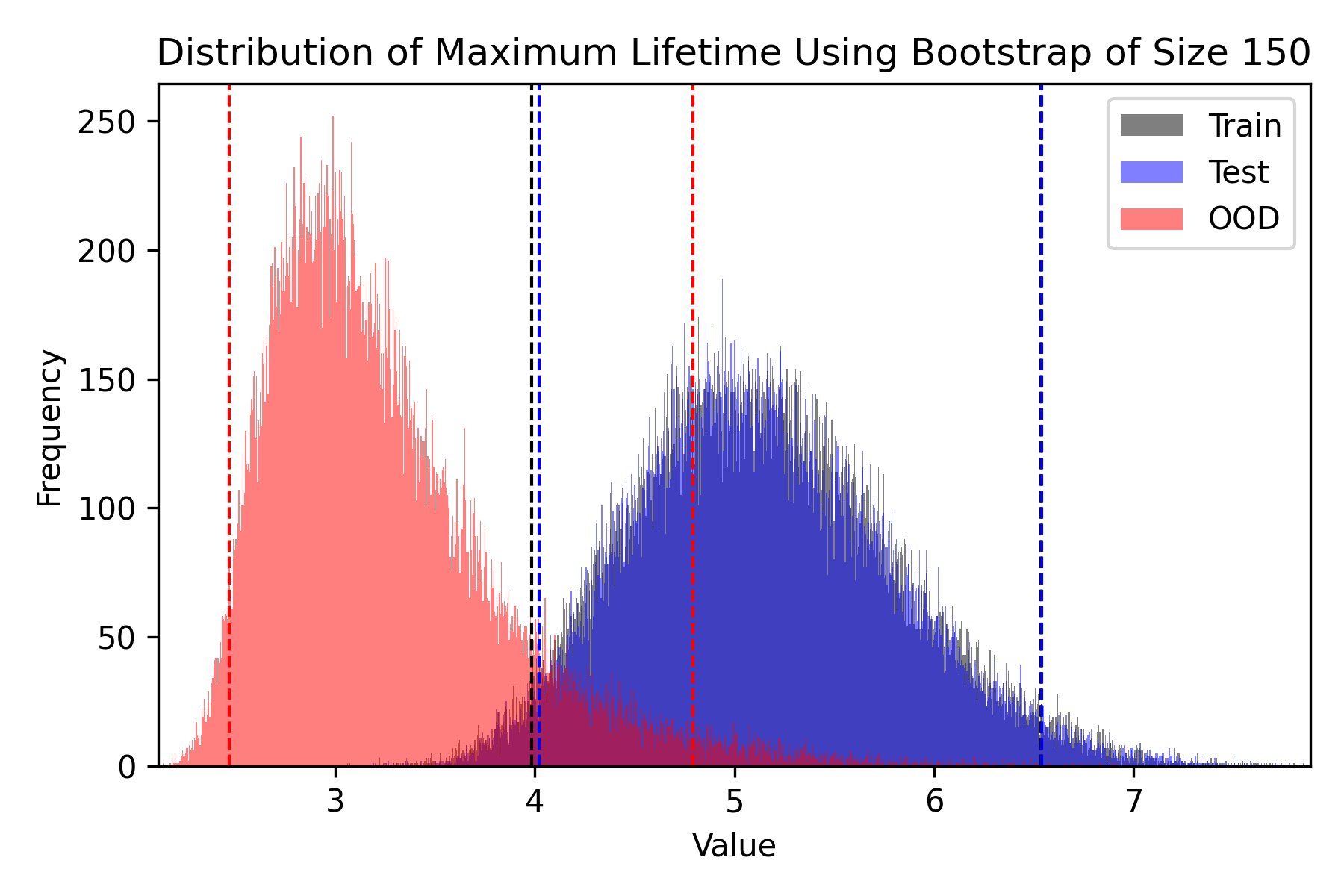}
    \end{subfigure}
    \vspace{-0.5cm}
    \caption{Distribution of average (left) and maximum (right) lifetime of connected components ($H_0$) in bootstrapped samples from the training and testing MNIST embeddings and EMNIST embeddings. Vertical lines indicate the 95\% confidence interval.}
    \label{fig:h0_mnist}
\end{figure}

The Train and Test distributions are extremely similar for MNIST, leading them to overlap nearly completely in Figure \ref{fig:h0_mnist}. While the Test distribution is new to the model, it is not unfamiliar and we would expect it to be ``trivialized.'' That is, we would expect the average lifetime for Test samples to be similar to Train samples. In the case of MNIST, the two distributions are nearly indistinguishable for both average and maximum lifetime in $H_0$.The effect of trivialization holds for all classes and both training and test splits. 

For the average lifetime distribution (Figure \ref{fig:h0_mnist} - left), the OOD and Train/Test distributions are significantly far apart. The OOD distribution has a greater average lifetime, although the spread of all three distributions is similar. This suggests that the ResNet18 model cannot trivialize the connected components in the OOD embeddings as effectively as it can for Test/Train embeddings. 

\begin{wraptable}{r}{5.5cm}
    \centering
    \caption{The 95\% confidence intervals for $H_0$ Average Lifetime for MNIST}
    \label{tab:MNIST_CI}
    \begin{tabular}{@{}l|l@{}}
    \toprule
    Train & (1.000, 1.118) \\ \midrule
    Test  & (0.996, 1.113) \\ \midrule
    OOD   & (1.303, 1.460) \\ \bottomrule
    \end{tabular}
\end{wraptable}

While average lifetime can capture the extent of ``trivialization'' in a neural network, the maximum lifetime instead characterizes the outliers. When we consider maximum lifetime (Figure \ref{fig:h0_mnist} - right), the differences among the three distributions are not as stark. All three distributions have considerably more spread and the 95\% confident intervals overlap. 
One might expect the ``trivialization'' principle to extend to the outliers as well, but we observe that the maximum lifetime is typically greater for the Train distribution compared to the OOD distribution. This suggests the presence of some extreme outliers in the training split, but not so many that it would skew the average lifetime of the samples to be higher.
    
\begin{figure}[h!]
    \centering
    \begin{subfigure}[b]{0.45\textwidth}
        \centering
        \includegraphics[width=\linewidth]{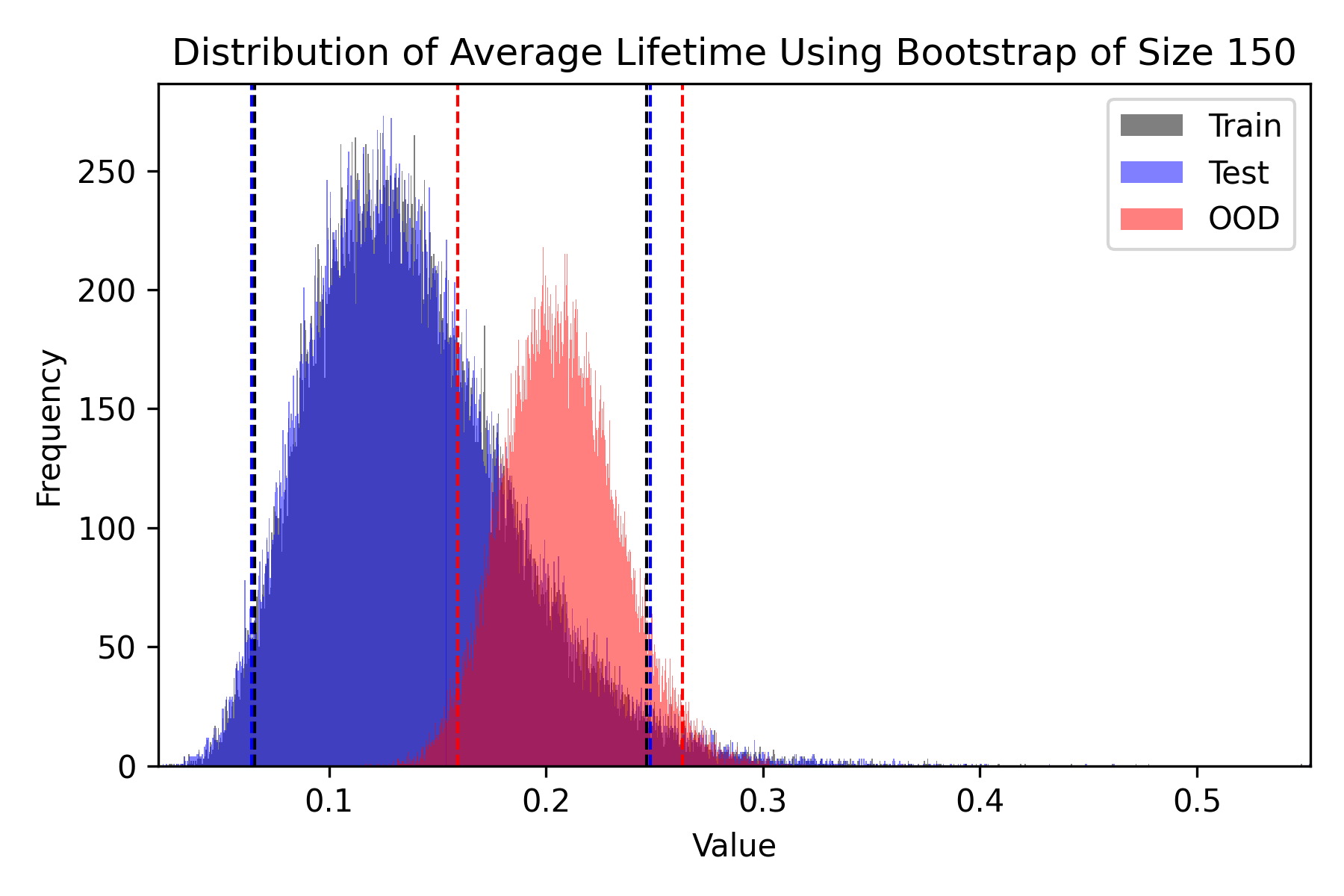}
    \end{subfigure}
    \hfill
    \begin{subfigure}[b]{0.45\textwidth}
        \centering
        \includegraphics[width=\linewidth]{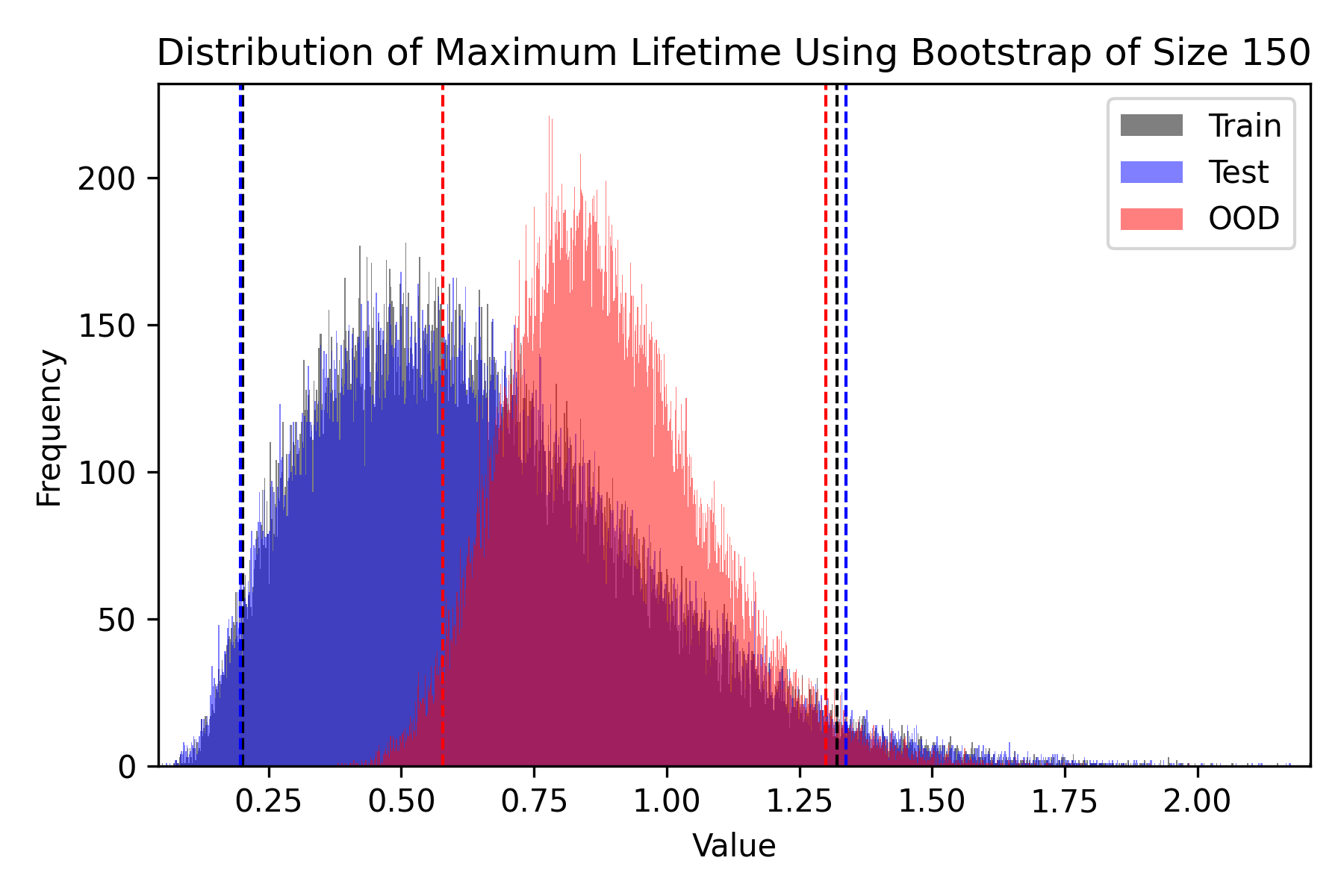}
    \end{subfigure}
    \vspace{-0.5cm}
    \caption{Distribution of average (left) and maximum (right) lifetime of connected components ($H_1$) in bootstrapped samples from the training and testing MNIST embeddings and EMNIST embeddings. Vertical lines indicate the 95\% confidence interval.}
    \label{fig:h1_mnist}
\end{figure}

Higher-dimensional persistence results can also be informative, but they resist the straightforward interpretation of ``proximity'' that can be ascribed to $H_0$. In $H_1$ (Figure \ref{fig:h1_mnist}), we investigate the presence of 1-dimensional topological ``holes'' and measure their average and maximum lifetimes. We see, again, a near-complete overlap of the Train and Test distributions. 
The average lifetime for all distributions is low, indicating that few higher-order features. The maximum lifetimes also resist interpretation. All three distributions show high spread and overlap. Generally, we may conclude that higher-order topological features do not persist in a well-trained model. Outliers are present, but the impact of such features overall is low.

\subsection{CIFAR-10}

\begin{figure}[htb]
    \centering
    \begin{subfigure}[b]{0.45\textwidth}
        \centering
        \includegraphics[width=\linewidth]{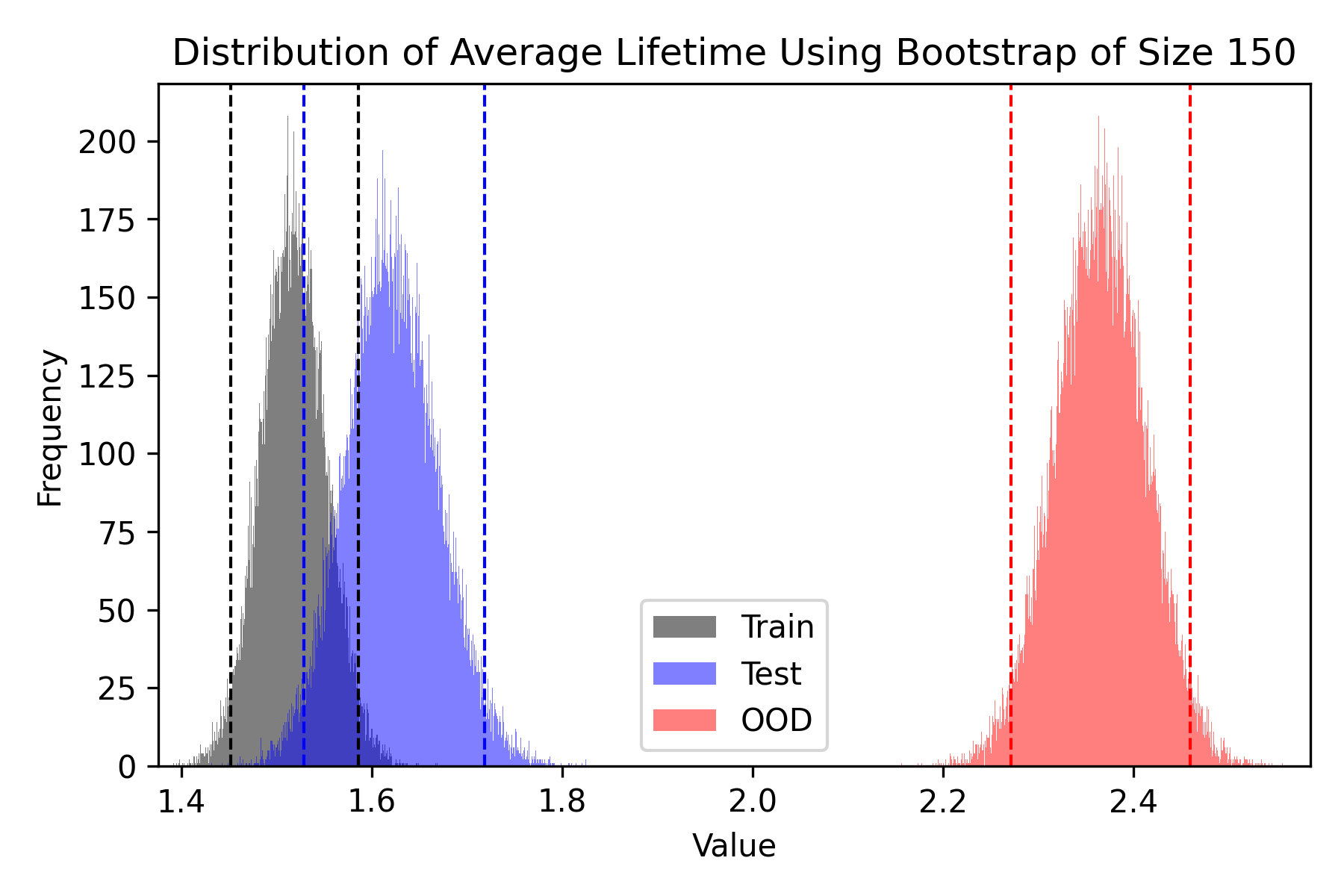}
        \label{fig:h0_avglifetime_cifar}
    \end{subfigure}
    \hfill
    \begin{subfigure}[b]{0.45\textwidth}
        \centering
        \includegraphics[width=\linewidth]{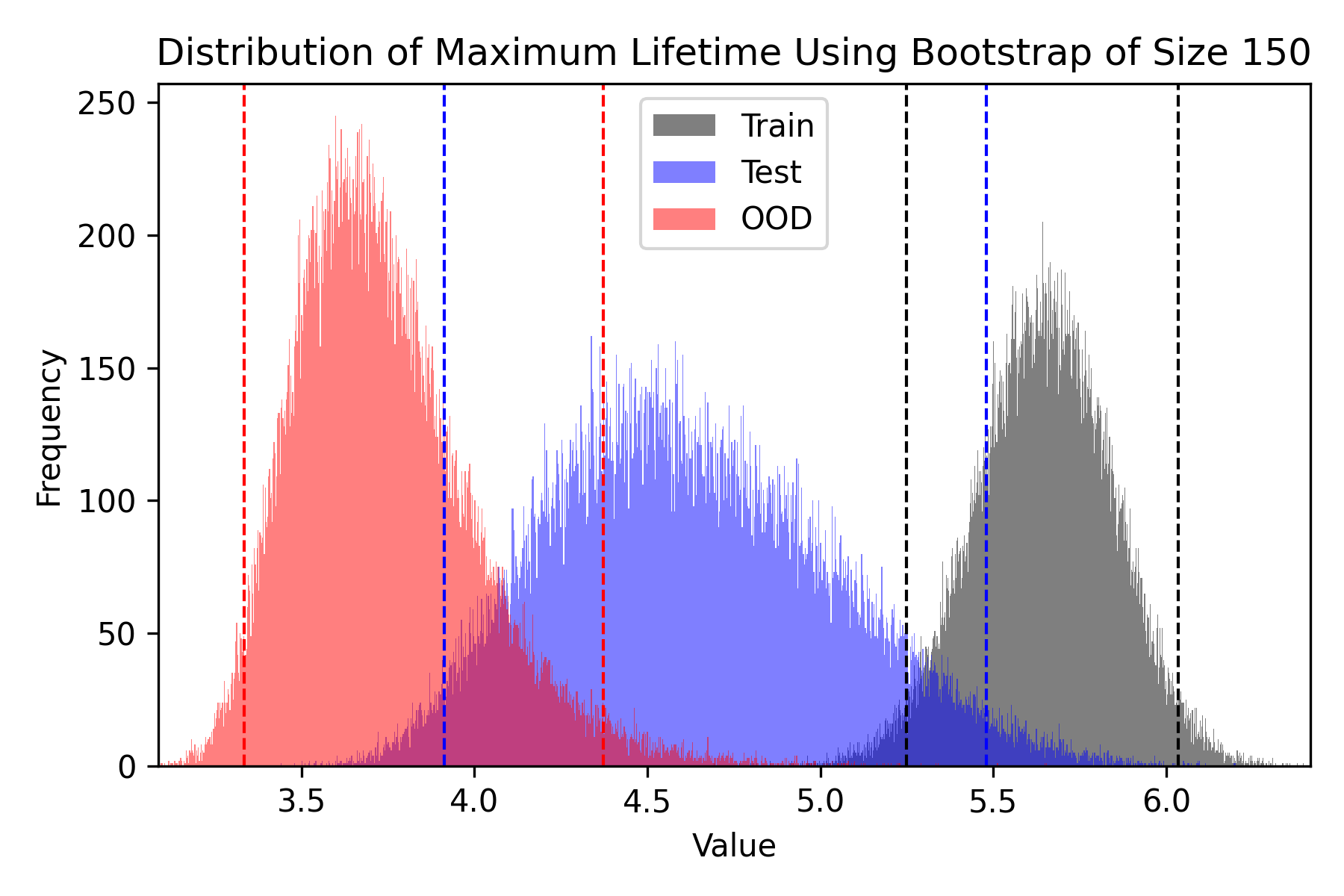}
        \label{fig:h0_maxlifetime_cifar}
    \end{subfigure}
    \vspace{-0.7cm}
    \caption{Distribution of average (left) and maximum (right) lifetime of connected components ($H_0$) in bootstrapped samples from the training and testing CIFAR-10 embeddings and CIFAR-100 embeddings. Vertical lines indicate the 95\% confidence interval.}
    \label{fig:h0_cifar}
\end{figure}

Figure \ref{fig:h0_cifar} shows the results of our bootstrapping approach on CIFAR-10 and CIFAR-100. We present the results for a bootstrap of size 150 (see \ref{sec:app} for other sample sizes). That is, the topological features measured are not for individual images in CIFAR-10 or CIFAR-100, but for batches of 150 randomly selected samples. Black and blue indicate the training and testing splits for CIFAR-10, respectively, while red indicates CIFAR-100. Recall that we perform the persistent homology computation for 50000 iterations for each distribution to account for inherent randomness. 

\begin{wraptable}{r}{5.5cm}
    \centering
    \caption{The 95\% confidence intervals for $H_0$ Average Lifetime for CIFAR}
    \label{tab:CIFAR_CI}
    \begin{tabular}{@{}l|l@{}}
    \toprule
    Train & (1.452, 1.586) \\ \midrule
    Test  & (1.529, 1.719) \\ \midrule
    OOD   & (2.272, 2.460) \\ \bottomrule
    \end{tabular}
\end{wraptable}

The average lifetime plot of Figure \ref{fig:h0_cifar} (left) shows the most dramatic difference among our datasets. We first note that the Train and OOD distributions are several standard deviations apart. On average, bootstrap samples of OOD examples have a longer persistence than the Training examples. This effect also holds when comparing the OOD distribution to the Test distribution (although the magnitude of the effect is marginally less). There is overlap both in the distribution and 95\% confidence intervals of the Test and Train datasets. The effect of ``trivialization'' holds across classes and across the testing and training split for CIFAR-10. 

Figure \ref{fig:h0_cifar} (right) shows the maximum lifetime of bootstrapped samples for our datasets. The distributions are not nearly as disjoint: OOD and Test intersect, as do Test and Train. Even the confidence intervals overlap by a significant margin. Notable too is the distinct shape of the distributions. The Test distribution has greater uncertainty compared to the other two. 

\begin{figure}[htb]
    \centering
    \begin{subfigure}[b]{0.45\textwidth}
        \centering
        \includegraphics[width=\linewidth]{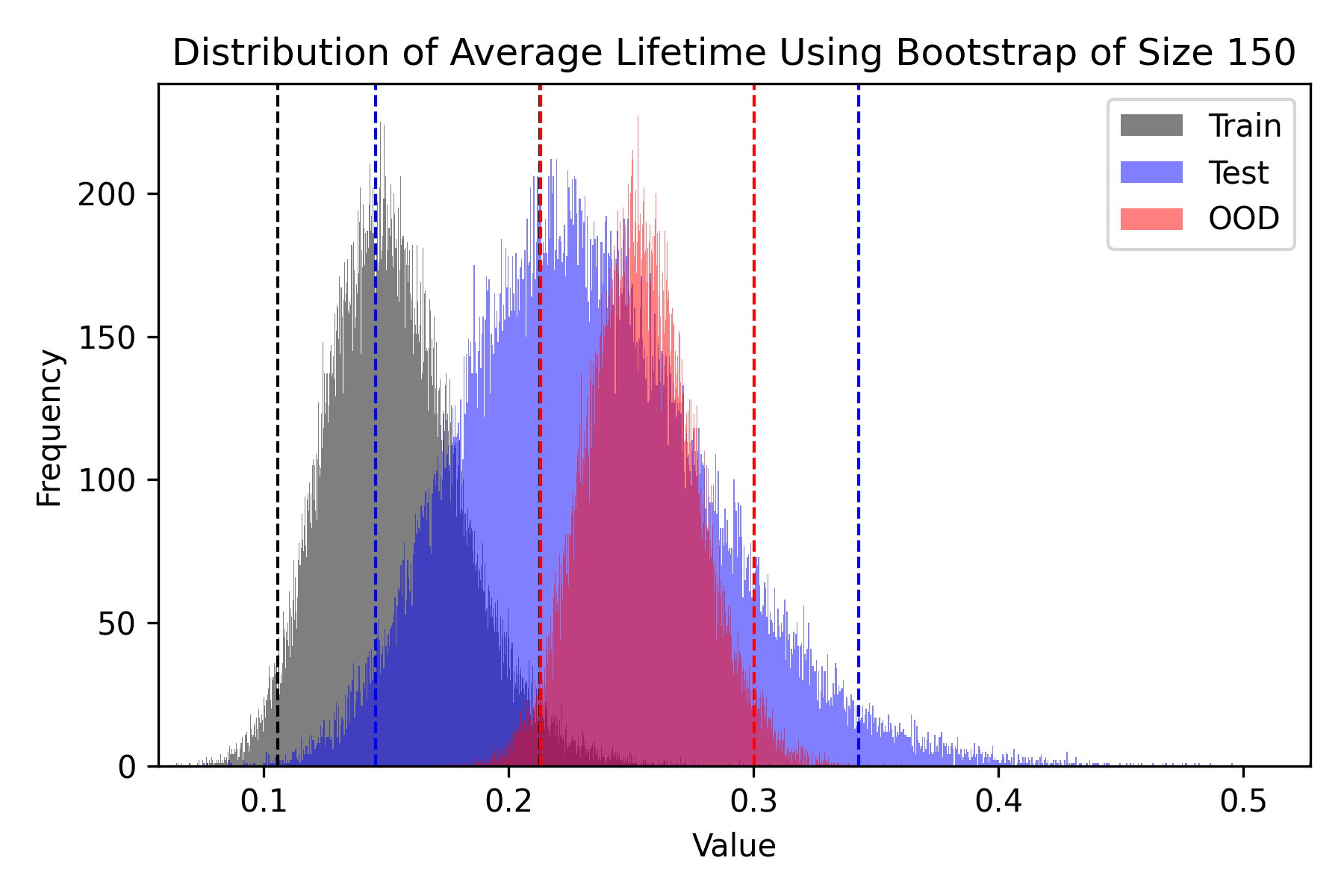}
        \label{fig:h1_avglifetime_cifar}
    \end{subfigure}
    \hfill
    \begin{subfigure}[b]{0.45\textwidth}
        \centering
        \includegraphics[width=\linewidth]{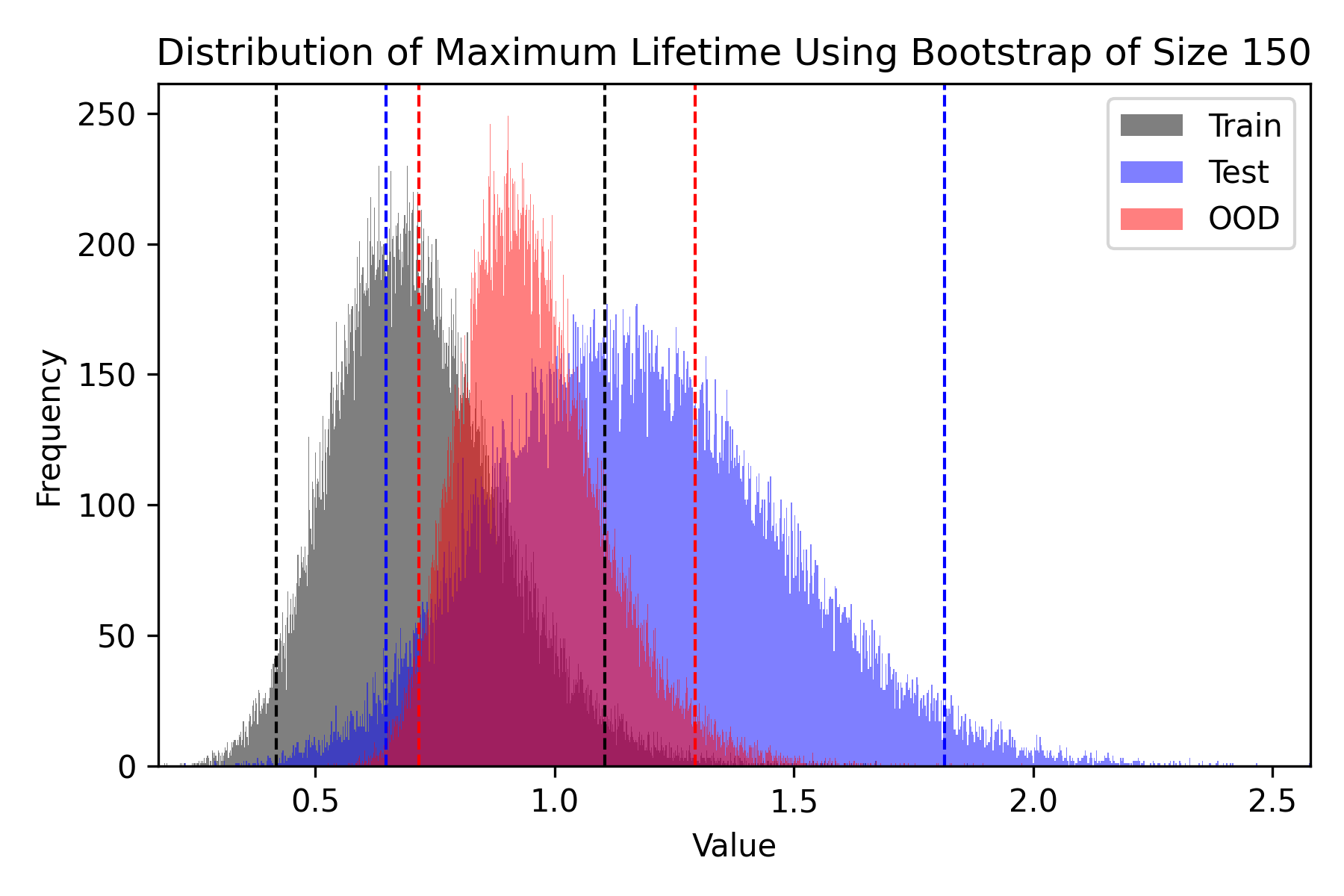}
        \label{fig:h1_maxlifetime_cifar}
    \end{subfigure}
    \vspace{-0.7cm}
    \caption{Distribution of average (left) and maximum (right) lifetime of topological ``holes'' ($H_1$) in bootstrapped samples from the training and testing CIFAR-10 embeddings and CIFAR-100 embeddings. Vertical lines indicate the 95\% confidence interval.}
    \label{fig:h1_cifar}
\end{figure}

The higher-order persistence $H_1$ results can be seen in Figure \ref{fig:h1_cifar}. There is still overlap across all three distributions, for both average and maximum lifetime, although the shape differs from that of the MNIST example (Figure \ref{fig:h1_mnist}). This may be due to the difference of complexity across the two examples. The CIFAR-10 embeddings appear to have low persistence for $H_1$ features, despite still containing some substantial outliers.

\section{Discussion}\label{sec:disc}

Our findings provide evidence that the topological simplification of data in DNNs holds in realistic, wide models and that this phenomenon can be leveraged for OOD detection. We demonstrate that there are detectable differences in the topology of ID and OOD examples. These differences are readily apparent in distributions of average lifetime in the lower-dimensional homology. Our experiments yield several crucial insights into DNNs that we highlight and discuss in this section: \begin{enumerate}
    \item topological simplification extends from training to test data; 
    \item the distance between distributions is indicative of dataset complexity; 
    \item meaningful topological features in DNNs occur at low dimensionality; and, 
    \item distributional differences in average lifetime can quantify uncertainty in models; 
\end{enumerate}

Previous works have focused on the topological simplification of training data. While our intuition suggests that this would hold for test data as well, there has not been (to our knowledge) similar empirical evidence of this phenomenon. Our work indicates that topological simplification, as indicated by the low average lifetime, occurs for realistic models across multiple datasets for the training and test splits. The average lifetime of each distribution may also be informative for a model's accuracy and generalizability. 

The distance between the average lifetime distributions are also indicative of dataset complexity. The MNIST dataset, which contains monochromatic images of handwritten digits, is far simpler than CIFAR-10, which comprises color images of animals and objects. It is well-known that accurately classifying MNIST is not sufficient evidence that a model is a strong image classifier. The fact that the training and test distributions for average lifetime are nearly identical in MNIST are suggestive of this simplicity. 

A model trained on MNIST can topologically simplify examples from the test split consistently, while a model trained on CIFAR-10 to the same degree of accuracy cannot do so to the same extent. We argue that this observation could be used to evaluate the complexity (relative to a model) of benchmark datasets in ML literature. Datasets whose latent layer embeddings yield a consistently higher average lifetime may be considered more challenging than those with lower average lifetime.  

Another insight of our experiments is the increasing computational feasibility of PH on latent layer embeddings of realistic DNNs. The need for TDA on wider, more representative models has been apparent in the literature, but often hampered by the computational cost of such an experiment. We find that the homology of latent layers can be studied using existing tools like Ripser with only small workarounds.

In our work, we find that bootstrap sampling allows us to process subsamples of high-dimensional data for tens of thousands of iterations, thereby providing a high-fidelity representation of the underlying manifold. We also find that one need not compute higher-dimensional homology. Our experiments uncovered stark topological differences using just $H_0$, which is by far the most inexpensive computation. This provides further evidence that TDA is computationally feasible for applications to realistic DNNs and that we can restrict ourselves to lower-dimensional topological features while still gleaning meaningful insights into our models.   

The motivating problem for this work was understanding OOD examples through the lens of TDA. Our experiments provide compelling evidence that topology can indicate model uncertainty for novel inputs. Namely, we can leverage the now well-known topological simplification that occurs in models to distinguish between ID and OOD data. Average lifetime in $H_0$ is a strong indicator for well-trained DNNs: low average lifetime is associated with ID examples, while high average lifetime is associated with OOD examples. We find this holds across multiple datasets of varying complexity. This observation informs a potential OOD detection pipeline, which uses subsampling of inputs to evaluate model uncertainty with TDA. 

One caveat to our  results is the sample size: as the dimension of structures increases, the number of datapoints needed to capture topological features does as well, as does the computation time. In Figure \ref{fig:h1_cifar}, it is difficult to determine a clear pattern differentiating the three distributions. OOD is centered about the highest average lifetime, but the $x$-axis indicates that the lifetimes were fairly short for all distributions. This may be a product of an insufficiently large sample size -- or it may be indicative of the ``trivialization.'' 

A well-trained model is expected to have few (if any) higher-order structures in the embedding space \cite{naitzat}, which is borne out in our results. Interestingly, even the OOD distribution indicates a low persistence of 1-dimensional features, which suggests that CIFAR-100 can still be trivialized to some extent. At the very least, the ResNet18 model is able to decrease the topological complexity of this higher-order feature, even though it cannot resolve $H_0$ features. 

This work opens up several avenues for future research. 

First and foremost, while our experiments indicate that topological invariants can identify OOD examples, it is crucial to test this for more architectures and datasets. Understanding the limitations of topological insights to DNNs is crucial for their usage in real-world situations. 
 
Further, we find that average and maximum lifetime may not be the best topological summary statistics for OOD detection. 

Alternative methods for comparing persistence diagrams, such as Betti curves or Wasserstein distance, should be considered. Additionally, PH is only the tip of the iceberg in regard to TDA methods. Exploration of OOD detection using topological tools like Mapper is a necessity.

\section*{Acknowledgements}
\textit{Sandia National Laboratories is a multi-mission laboratory managed and operated by National Technology and Engineering Solutions of Sandia, LLC., a wholly owned subsidiary of Honeywell International, Inc., for the U.S. Department of Energy's National Nuclear Security Administration under contract DE-NA-0003525. This paper describes objective technical results and analysis. Any subjective views or opinions that might be expressed in the paper do not necessarily represent the views of the U.S. Department of Energy or the United States Government.}


\vskip 0.2in
\bibliographystyle{abbrvnat}
\bibliography{jmlr-bib}

\appendix
\section{Additional Figures}\label{sec:app}

For each dataset, we tested multiple sample sizes to ensure that our bootstrap size was sufficiently large. As one may expect, the $H_0$ distributions for each dataset become more normal with smaller standard deviation as the sample size increases. We were constrained by compute time, as too large of a sample causes prohibitively expensive $H_1$ computations. 

\subsection{MNIST Figures}
\begin{figure}[ht]
    \centering
    \begin{subfigure}[b]{0.45\textwidth}
        \centering
        \includegraphics[width=\linewidth]{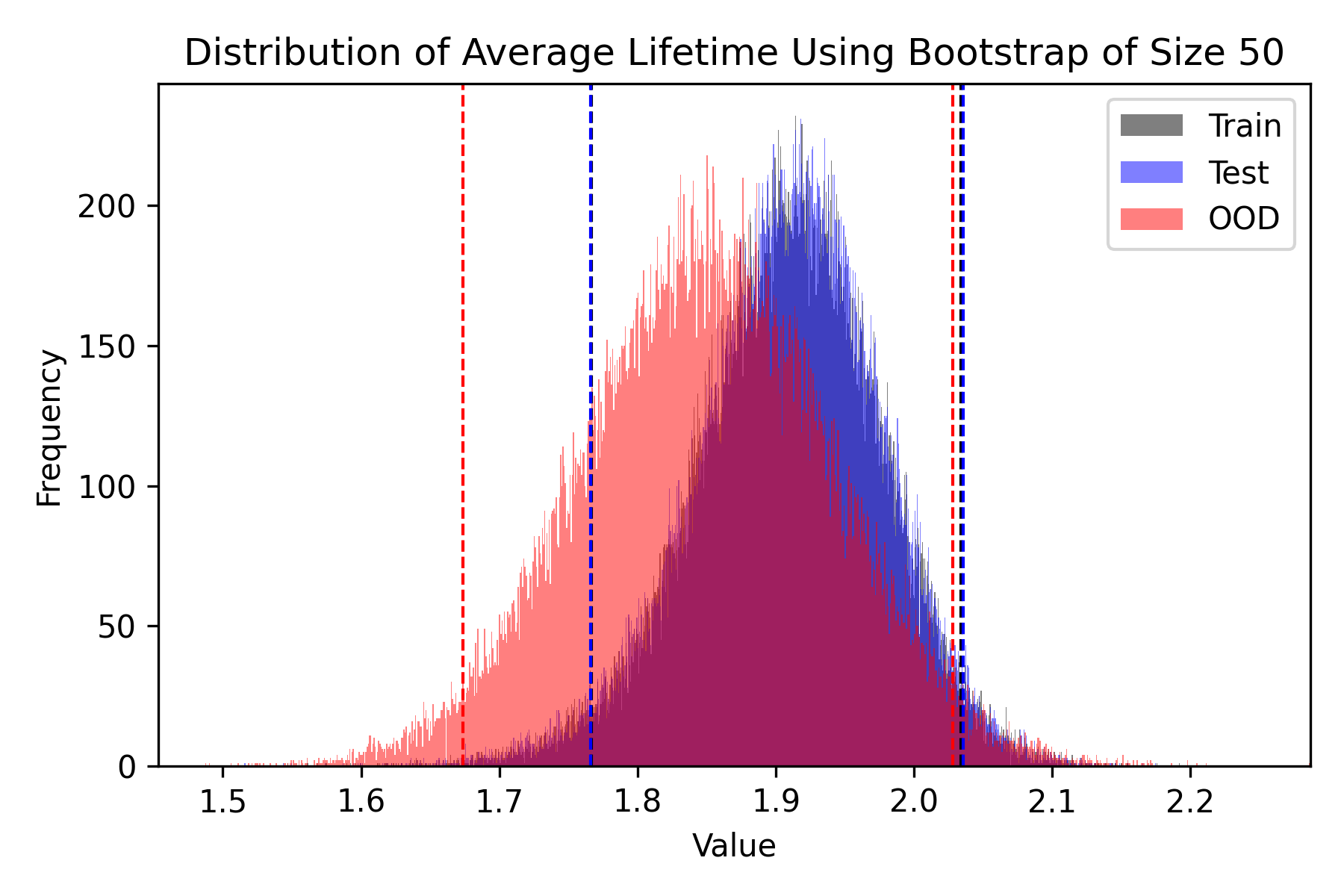}
    \end{subfigure}
    \hfill
    \begin{subfigure}[b]{0.45\textwidth}
        \centering
        \includegraphics[width=\linewidth]{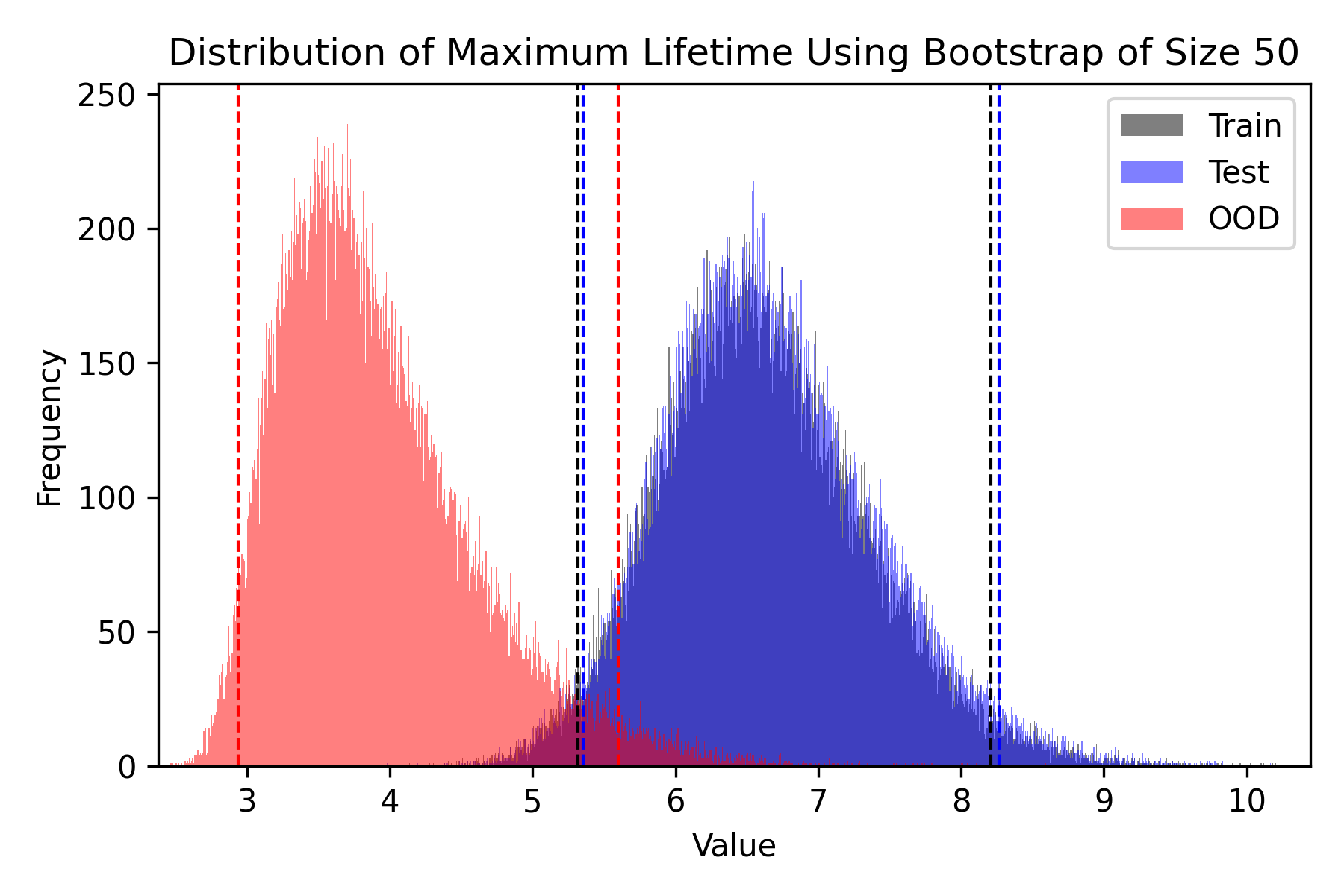}
    \end{subfigure}
    \vspace{-0.5cm}
    \caption{Distribution of average (left) and maximum (right) lifetime of connected components ($H_0$) in bootstrapped samples of size 50 from the training and testing MNIST embeddings and EMNIST embeddings. Vertical lines indicate the 95\% confidence interval.}
\end{figure}

\begin{figure}[h!]
    \centering
    \begin{subfigure}[b]{0.45\textwidth}
        \centering
        \includegraphics[width=\linewidth]{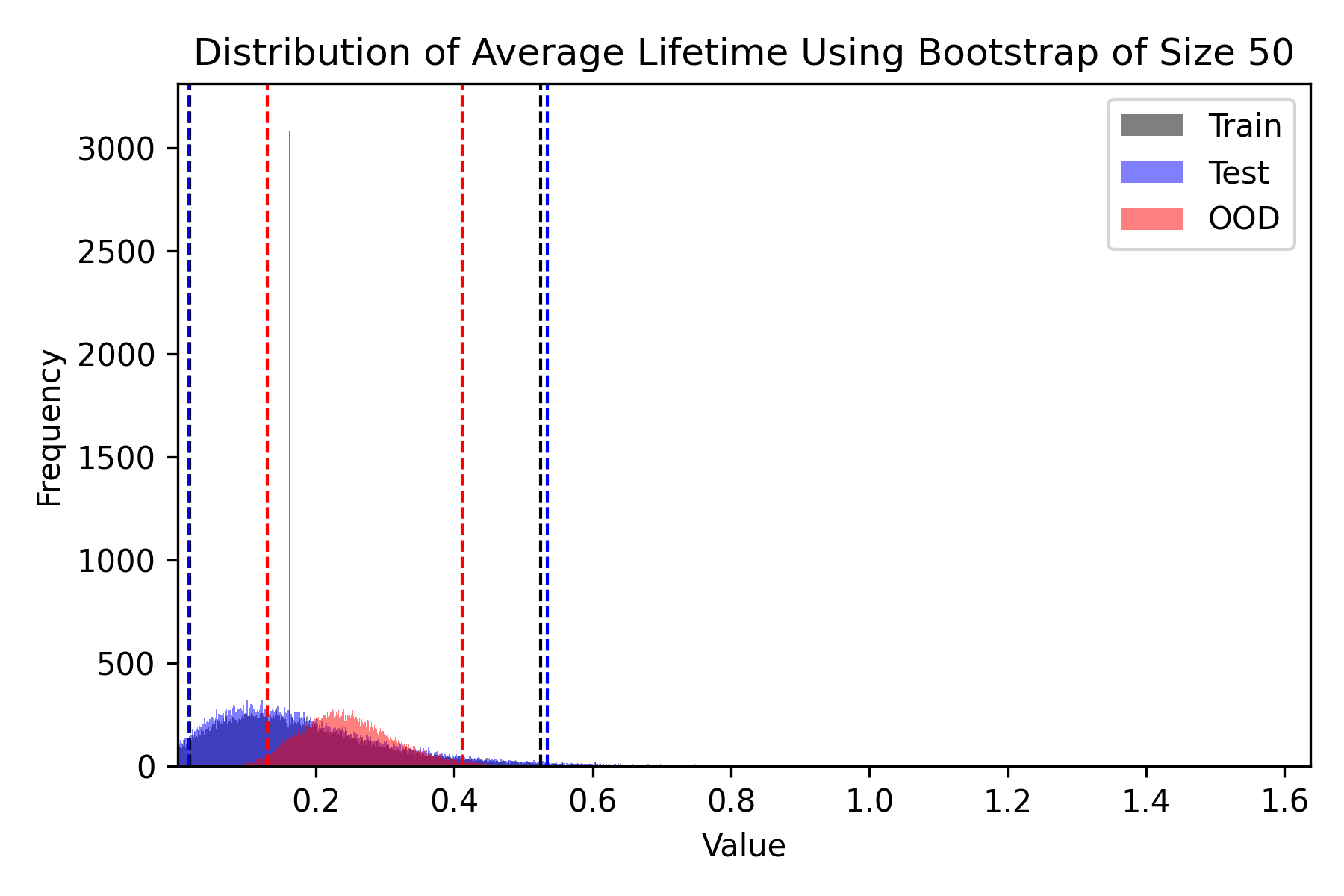}
    \end{subfigure}
    \hfill
    \begin{subfigure}[b]{0.45\textwidth}
        \centering
        \includegraphics[width=\linewidth]{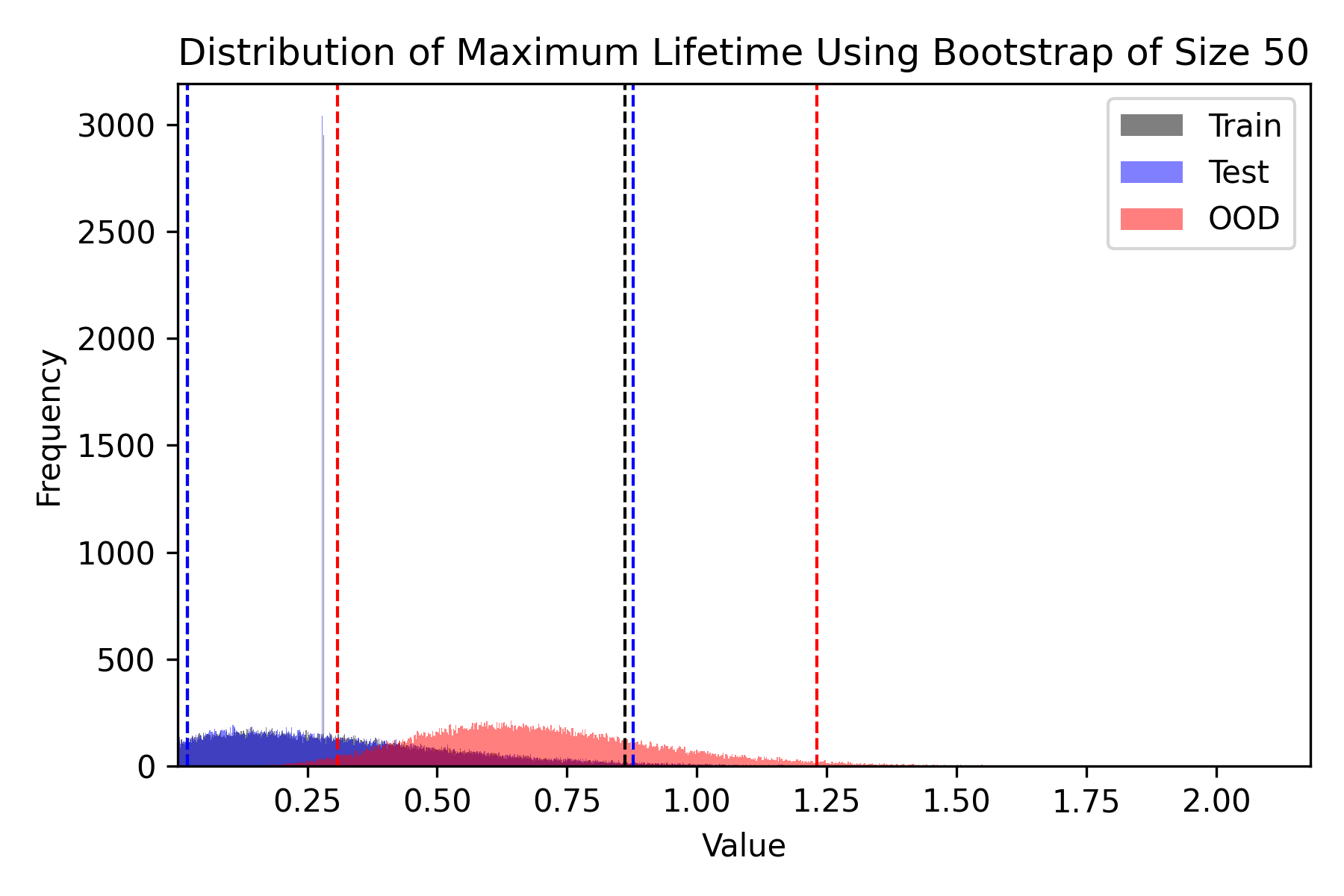}
    \end{subfigure}
    \vspace{-0.5cm}
    \caption{Distribution of average (left) and maximum (right) lifetime of holes ($H_1$) in bootstrapped samples of size 50 from the training and testing MNIST embeddings and EMNIST embeddings. Vertical lines indicate the 95\% confidence interval.}
\end{figure}

\begin{figure}[h!]
    \centering
    \begin{subfigure}[b]{0.45\textwidth}
        \centering
        \includegraphics[width=\linewidth]{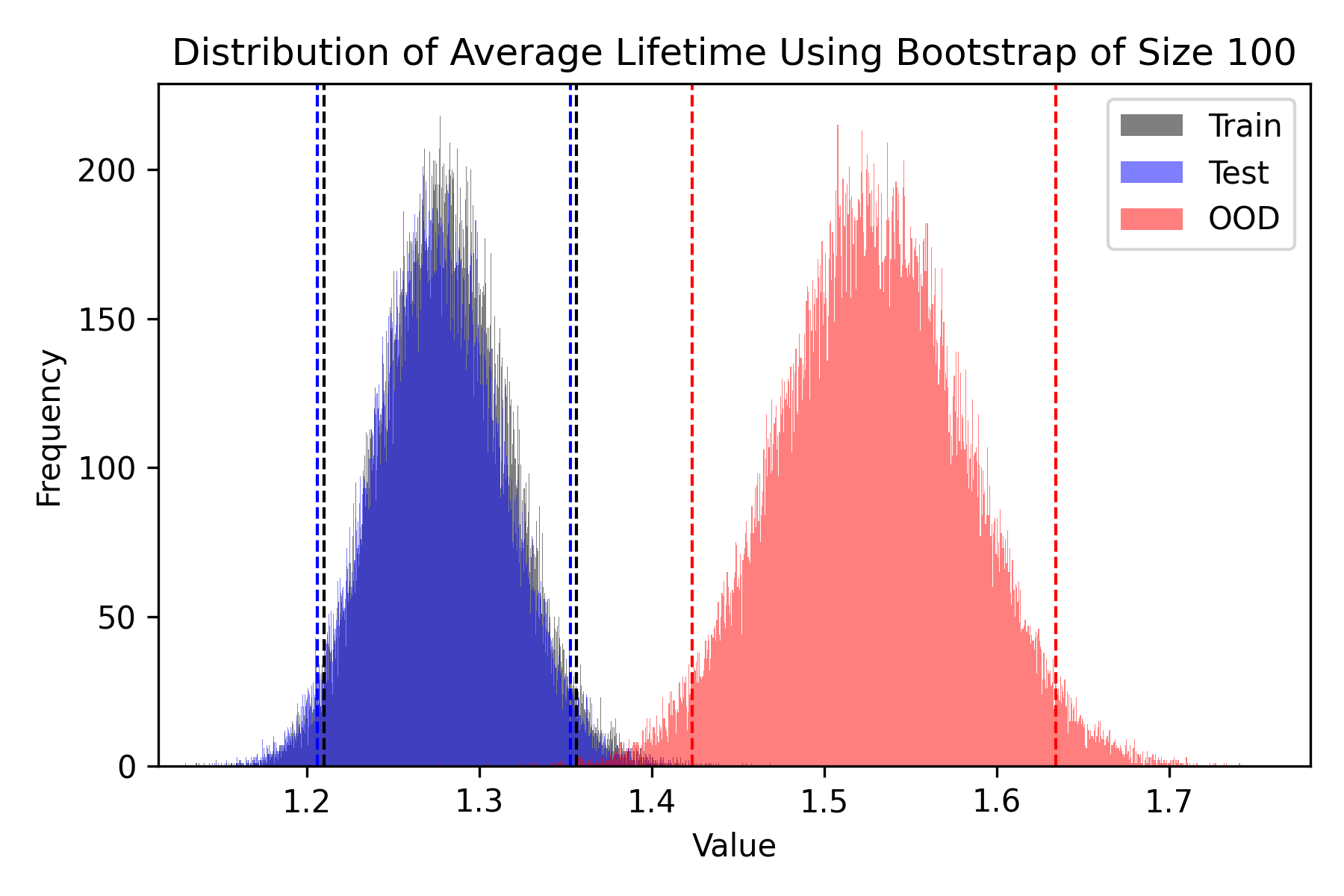}
    \end{subfigure}
    \hfill
    \begin{subfigure}[b]{0.45\textwidth}
        \centering
        \includegraphics[width=\linewidth]{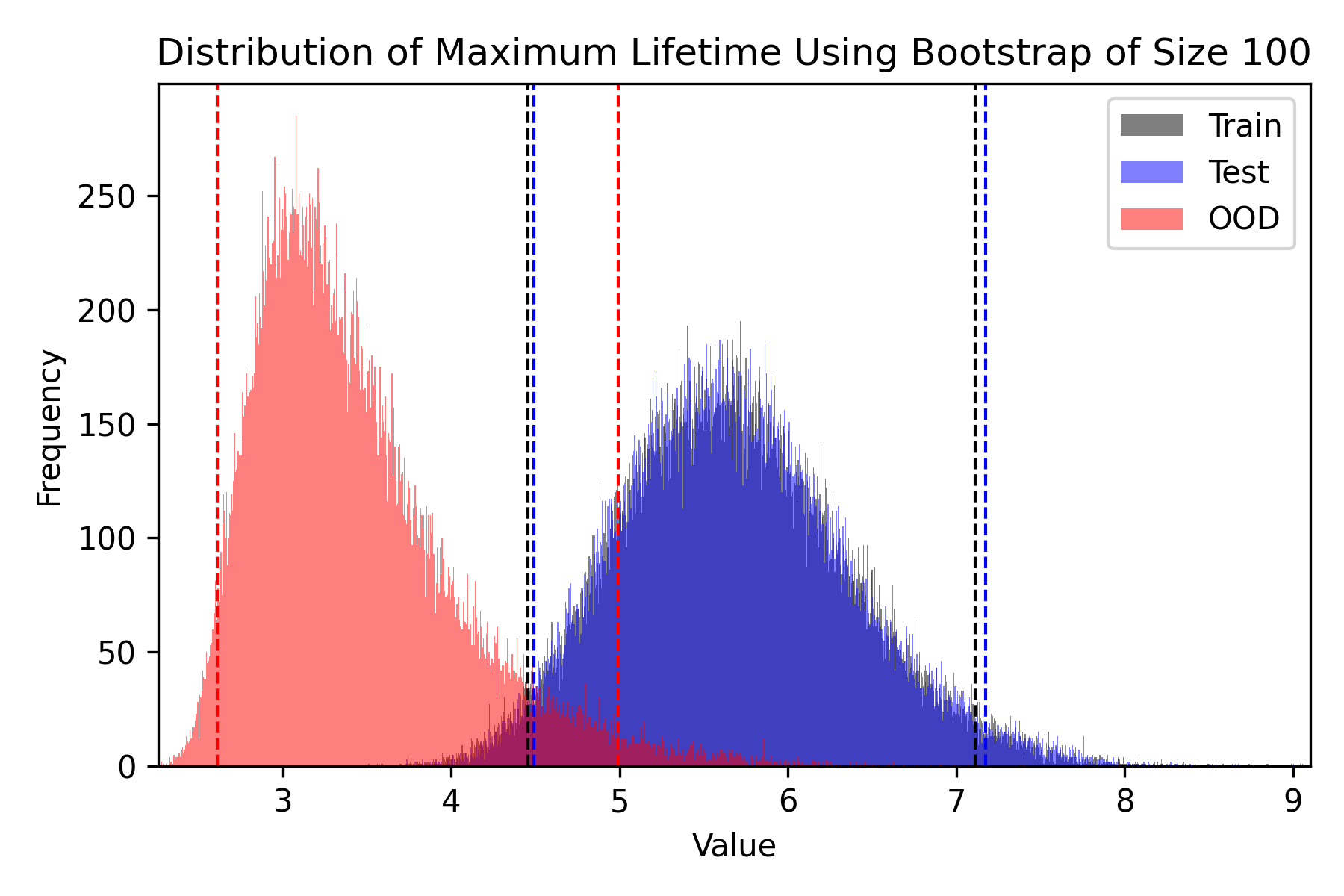}
    \end{subfigure}
    \vspace{-0.5cm}
    \caption{Distribution of average (left) and maximum (right) lifetime of connected components ($H_0$) in bootstrapped samples of size 100 from the training and testing MNIST embeddings and EMNIST embeddings. Vertical lines indicate the 95\% confidence interval.}
\end{figure}

\begin{figure}[h!]
    \centering
    \begin{subfigure}[b]{0.45\textwidth}
        \centering
        \includegraphics[width=\linewidth]{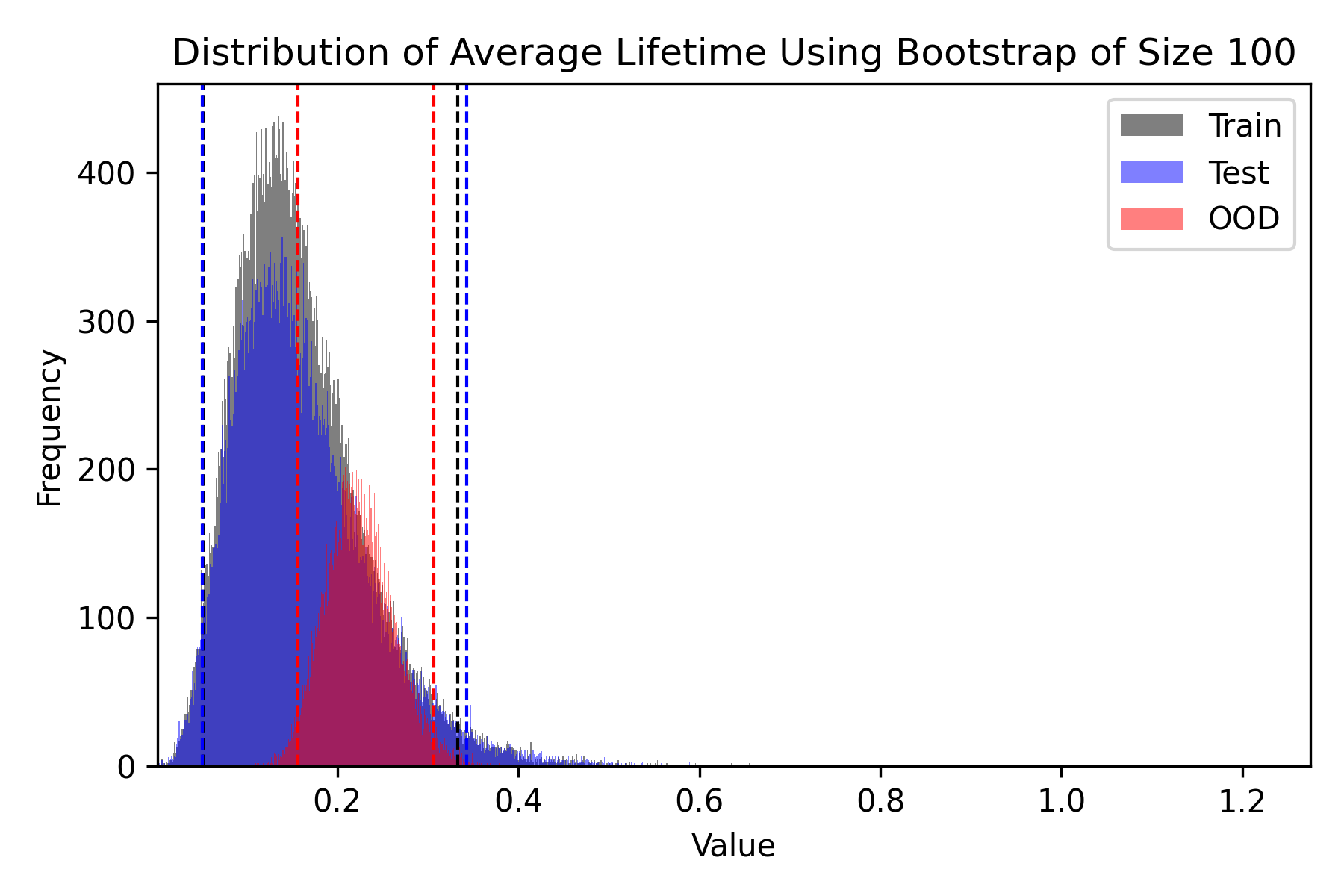}
    \end{subfigure}
    \hfill
    \begin{subfigure}[b]{0.45\textwidth}
        \centering
        \includegraphics[width=\linewidth]{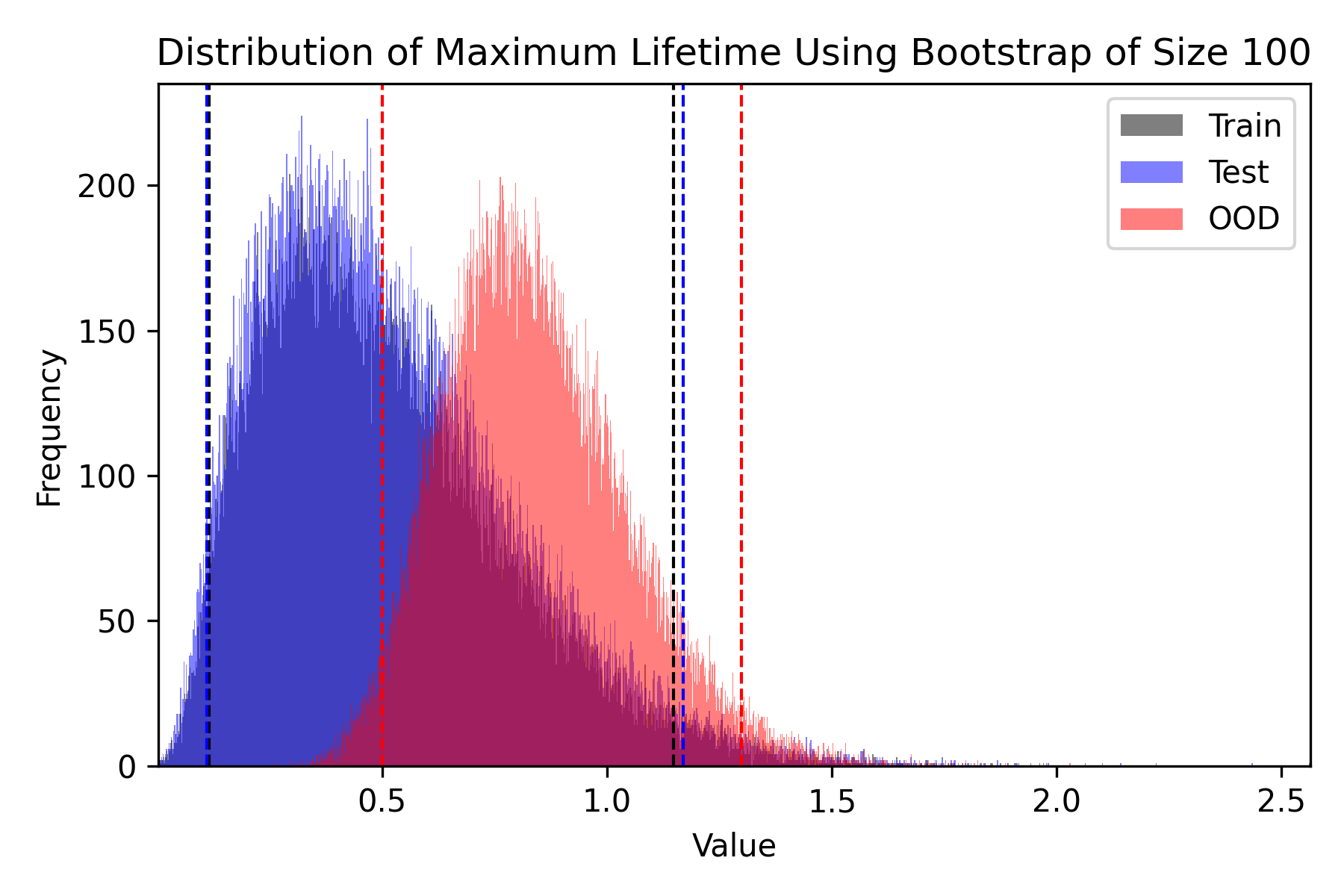}
    \end{subfigure}
    \vspace{-0.5cm}
    \caption{Distribution of average (left) and maximum (right) lifetime of holes ($H_1$) in bootstrapped samples of size 100 from the training and testing MNIST embeddings and EMNIST embeddings. Vertical lines indicate the 95\% confidence interval.}
\end{figure}

\newpage
\subsection{CIFAR Figures}

\begin{figure}[h!]
    \centering
    \begin{subfigure}[b]{0.45\textwidth}
        \centering
        \includegraphics[width=\linewidth]{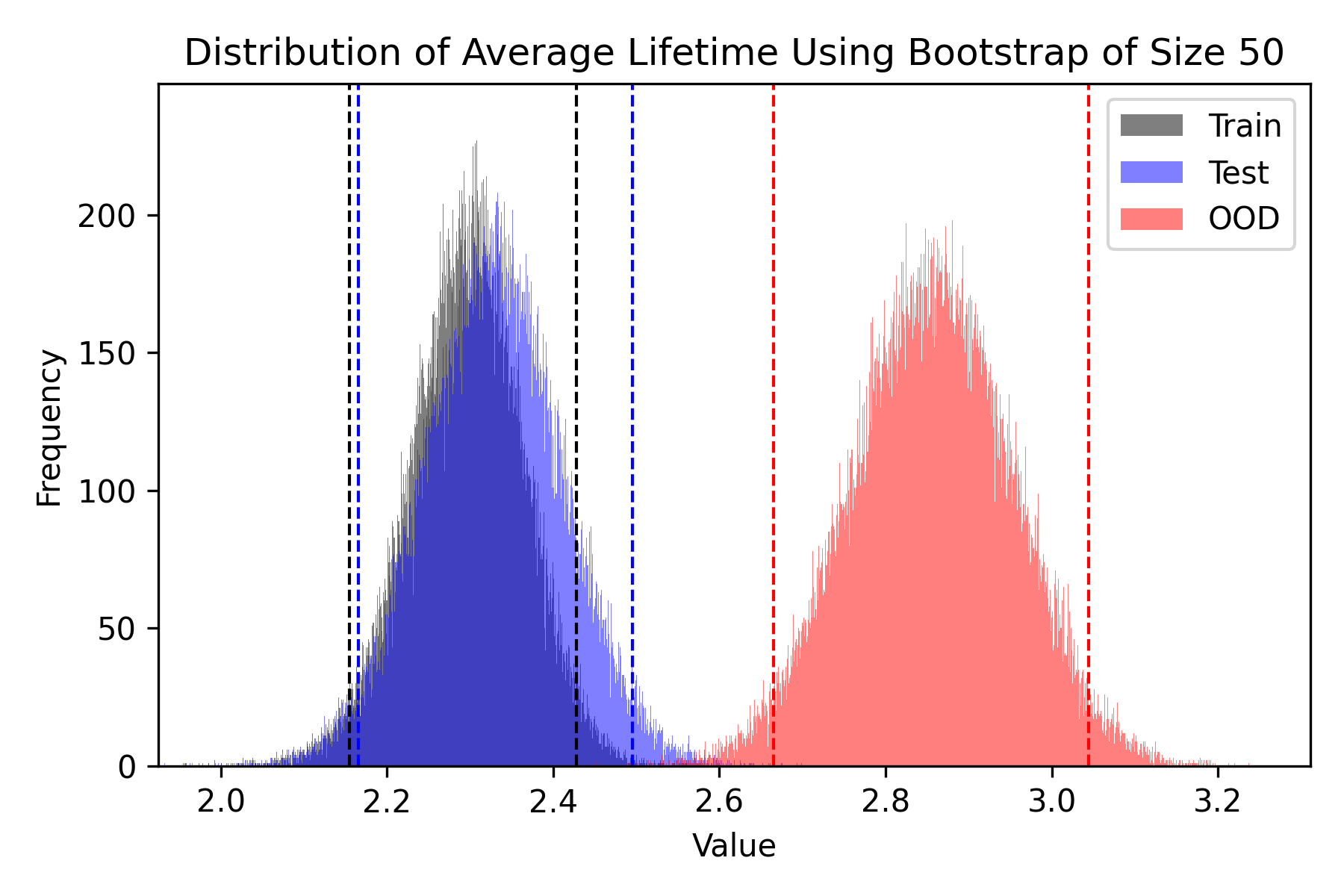}
    \end{subfigure}
    \hfill
    \begin{subfigure}[b]{0.45\textwidth}
        \centering
        \includegraphics[width=\linewidth]{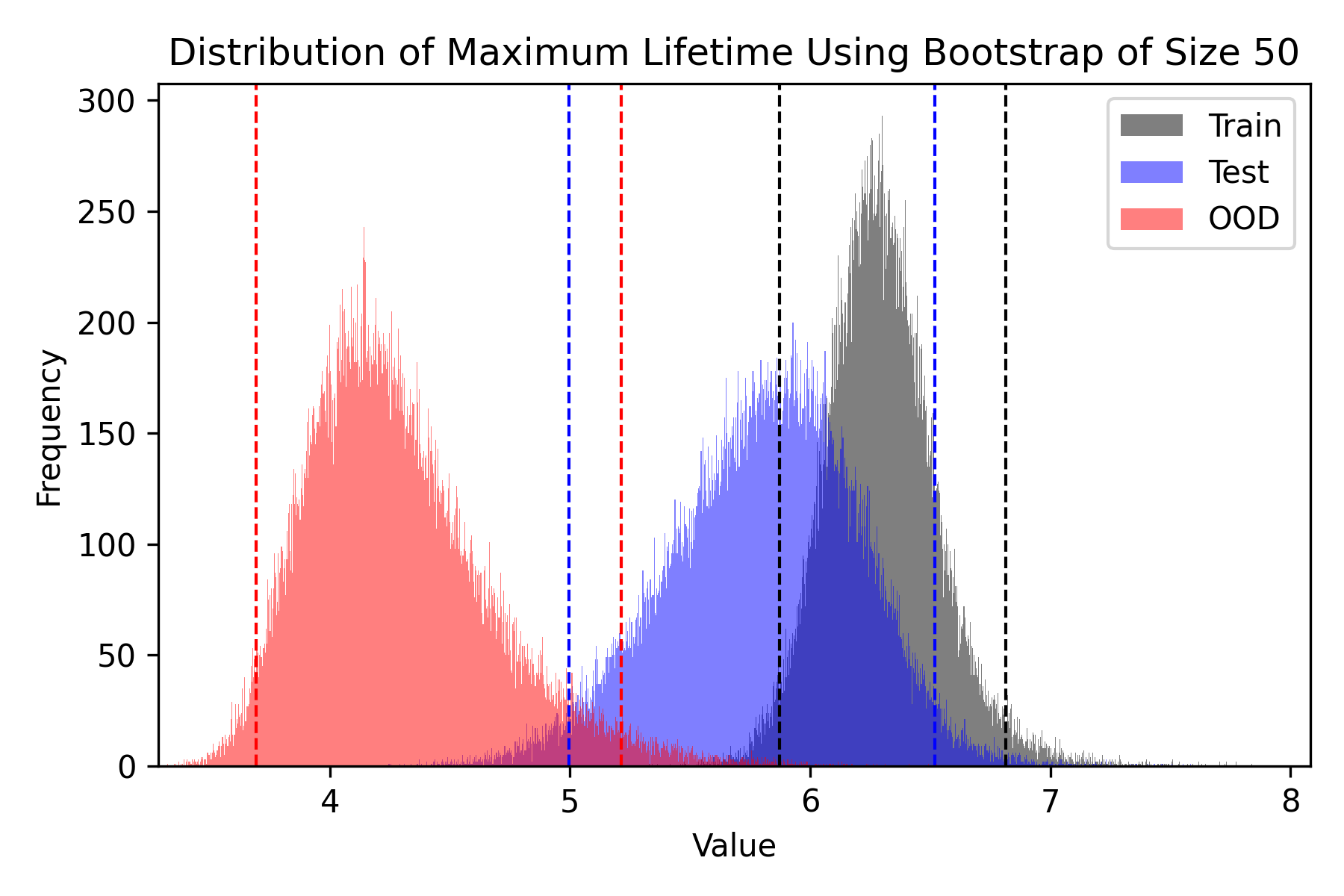}
    \end{subfigure}
    \vspace{-0.5cm}
    \caption{Distribution of average (left) and maximum (right) lifetime of connected components ($H_0$) in bootstrapped samples of size 50 from the training and testing CIFAR-10 embeddings and CIFAR-100 embeddings. Vertical lines indicate the 95\% confidence interval.}
\end{figure}

\begin{figure}[h!]
    \centering
    \begin{subfigure}[b]{0.45\textwidth}
        \centering
        \includegraphics[width=\linewidth]{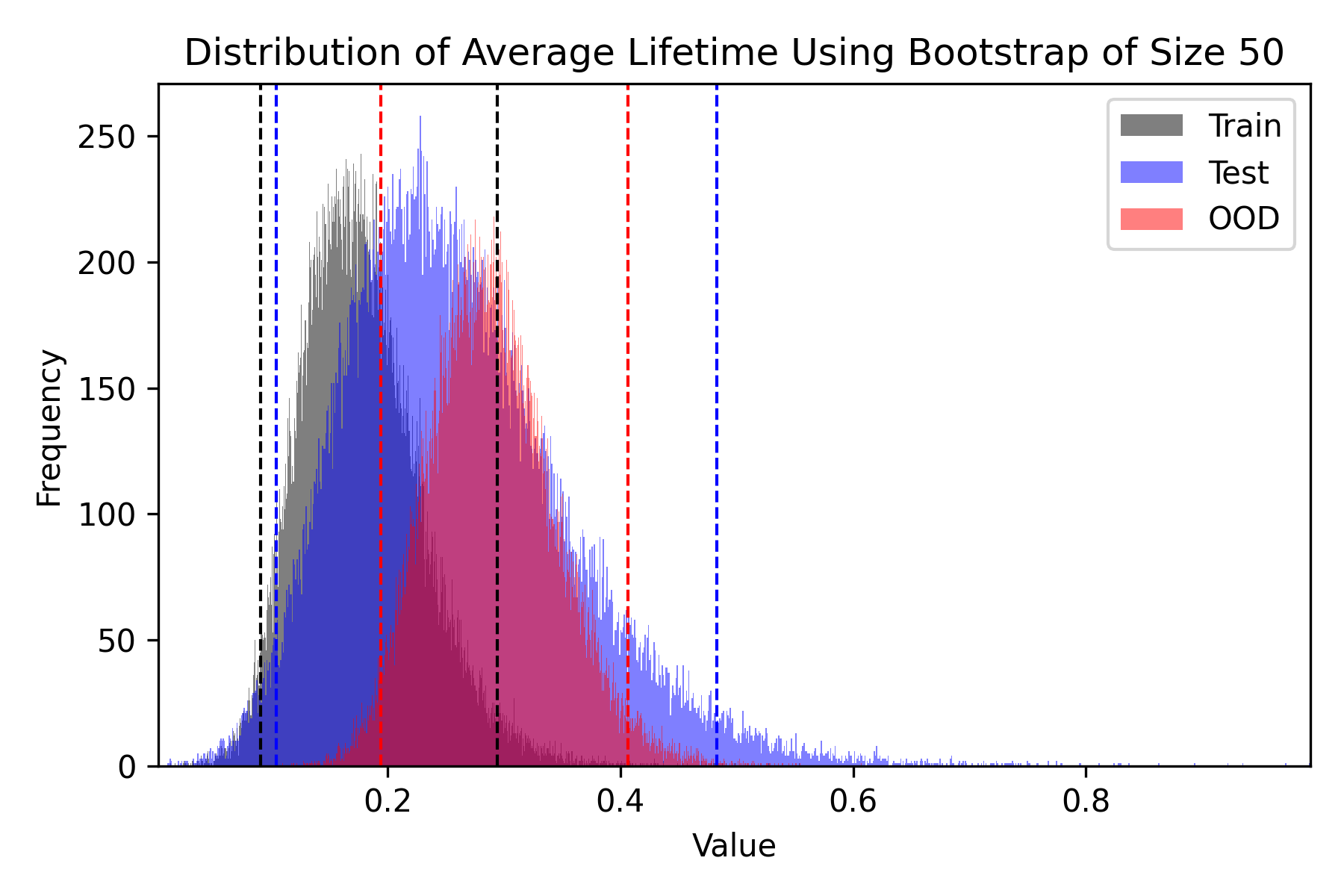}
    \end{subfigure}
    \hfill
    \begin{subfigure}[b]{0.45\textwidth}
        \centering
        \includegraphics[width=\linewidth]{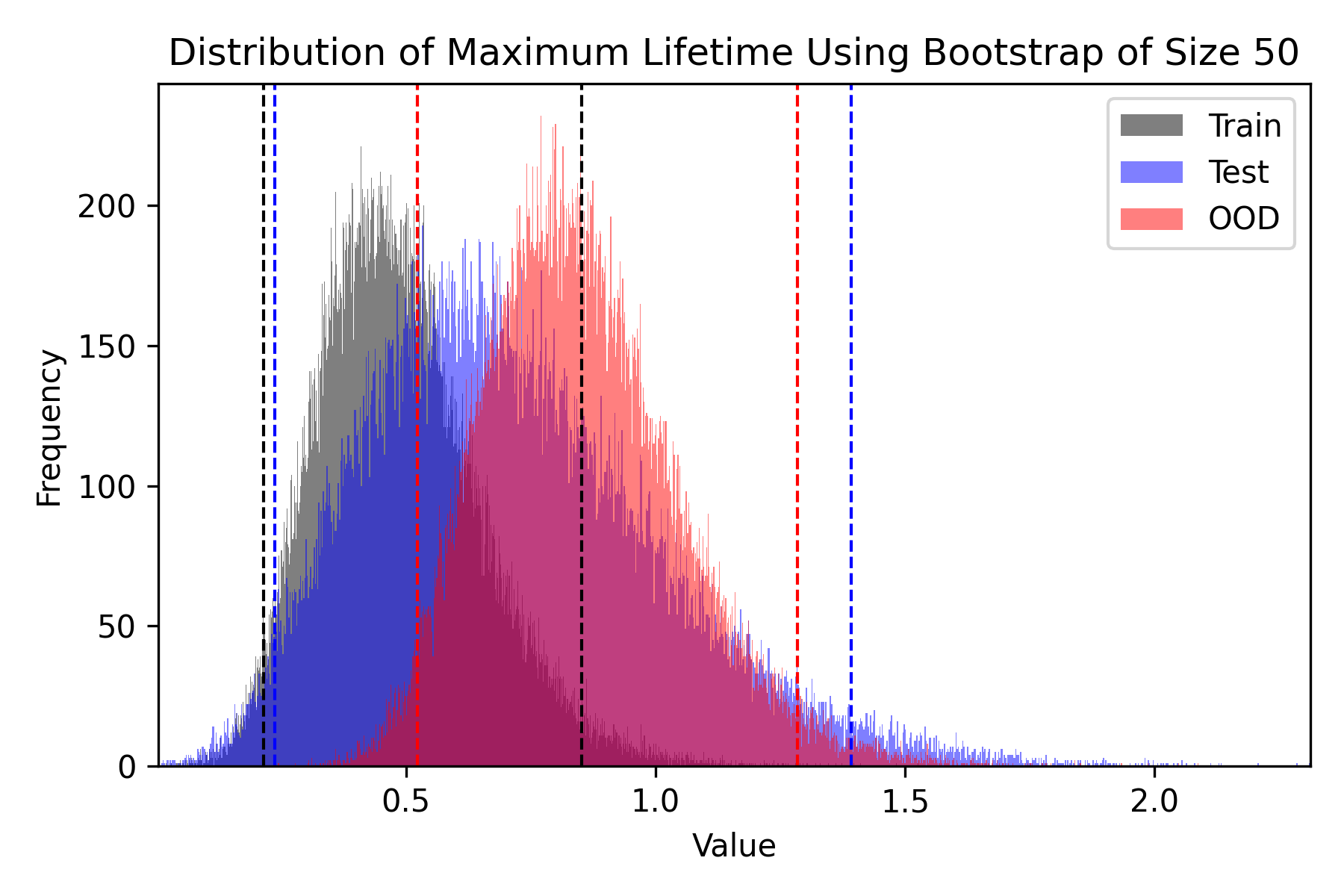}
    \end{subfigure}
    \vspace{-0.5cm}
    \caption{Distribution of average (left) and maximum (right) lifetime of holes ($H_1$) in bootstrapped samples of size 50 from the training and testing CIFAR-10 embeddings and CIFAR-100 embeddings. Vertical lines indicate the 95\% confidence interval.}
\end{figure}

\begin{figure}[h!]
    \centering
    \begin{subfigure}[b]{0.45\textwidth}
        \centering
        \includegraphics[width=\linewidth]{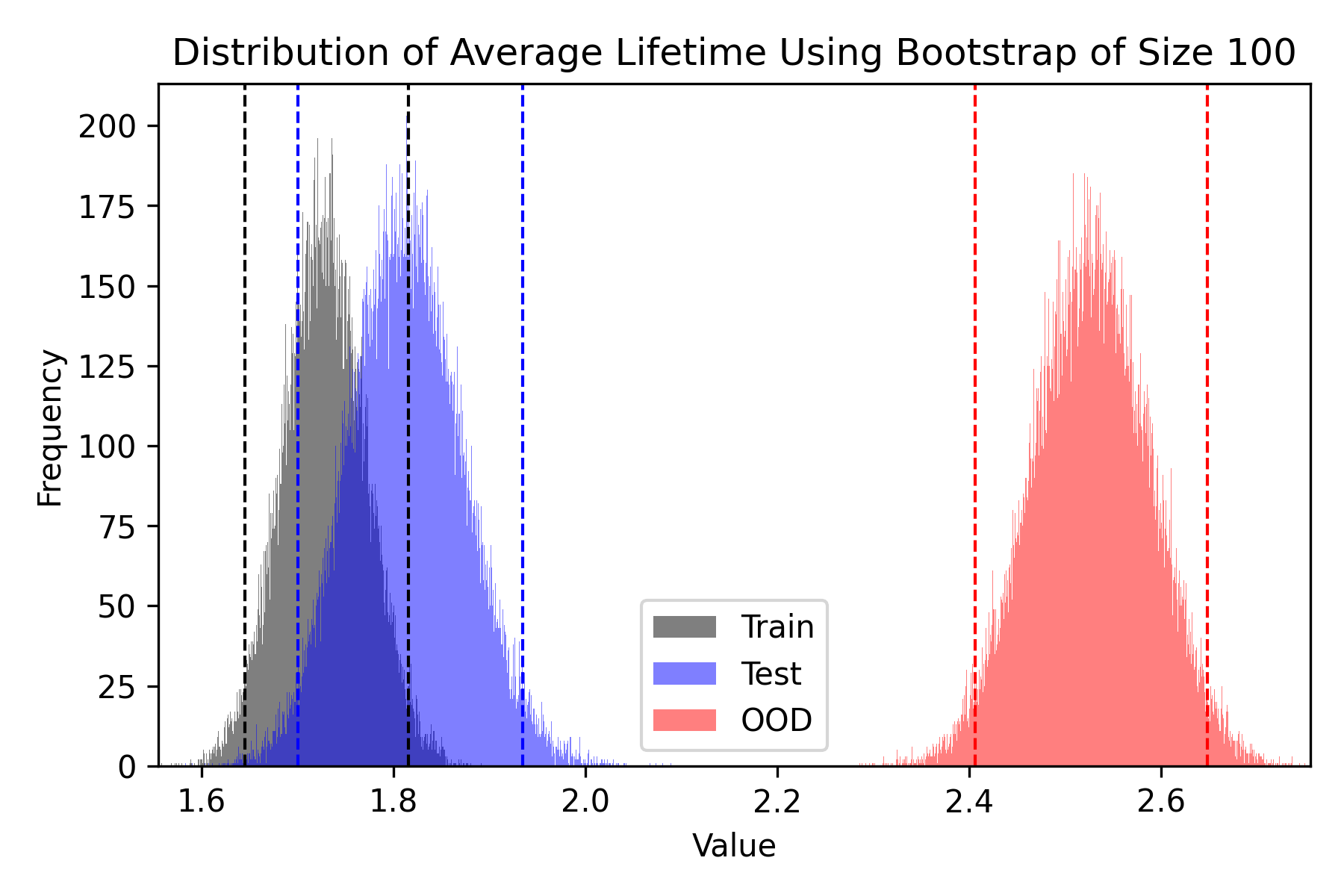}
    \end{subfigure}
    \hfill
    \begin{subfigure}[b]{0.45\textwidth}
        \centering
        \includegraphics[width=\linewidth]{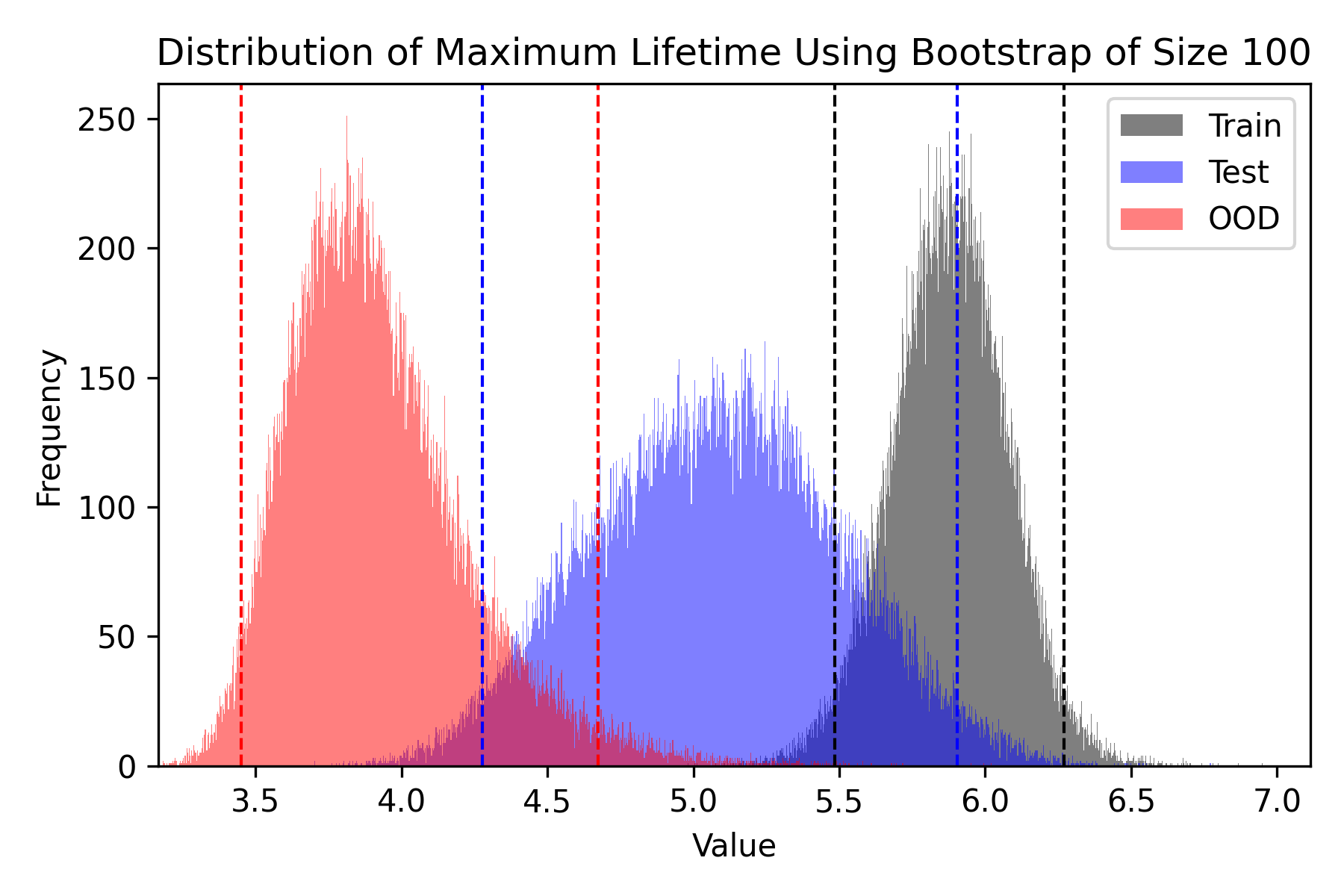}
    \end{subfigure}
    \vspace{-0.5cm}
    \caption{Distribution of average (left) and maximum (right) lifetime of connected components ($H_0$) in bootstrapped samples of size 100 from the training and testing CIFAR-10 embeddings and CIFAR-100 embeddings. Vertical lines indicate the 95\% confidence interval.}
\end{figure}

\begin{figure}[h!]
    \centering
    \begin{subfigure}[b]{0.45\textwidth}
        \centering
        \includegraphics[width=\linewidth]{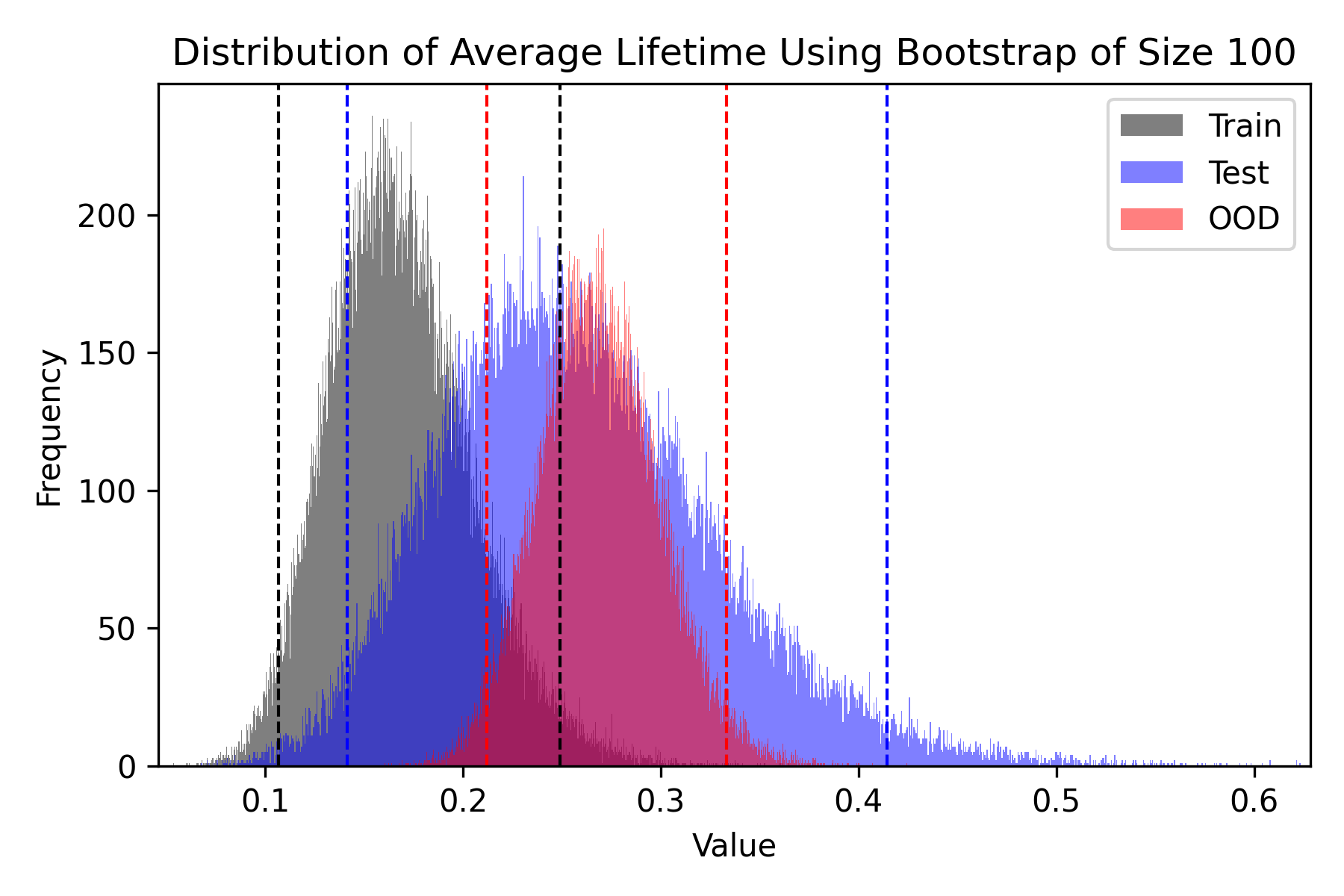}
    \end{subfigure}
    \hfill
    \begin{subfigure}[b]{0.45\textwidth}
        \centering
        \includegraphics[width=\linewidth]{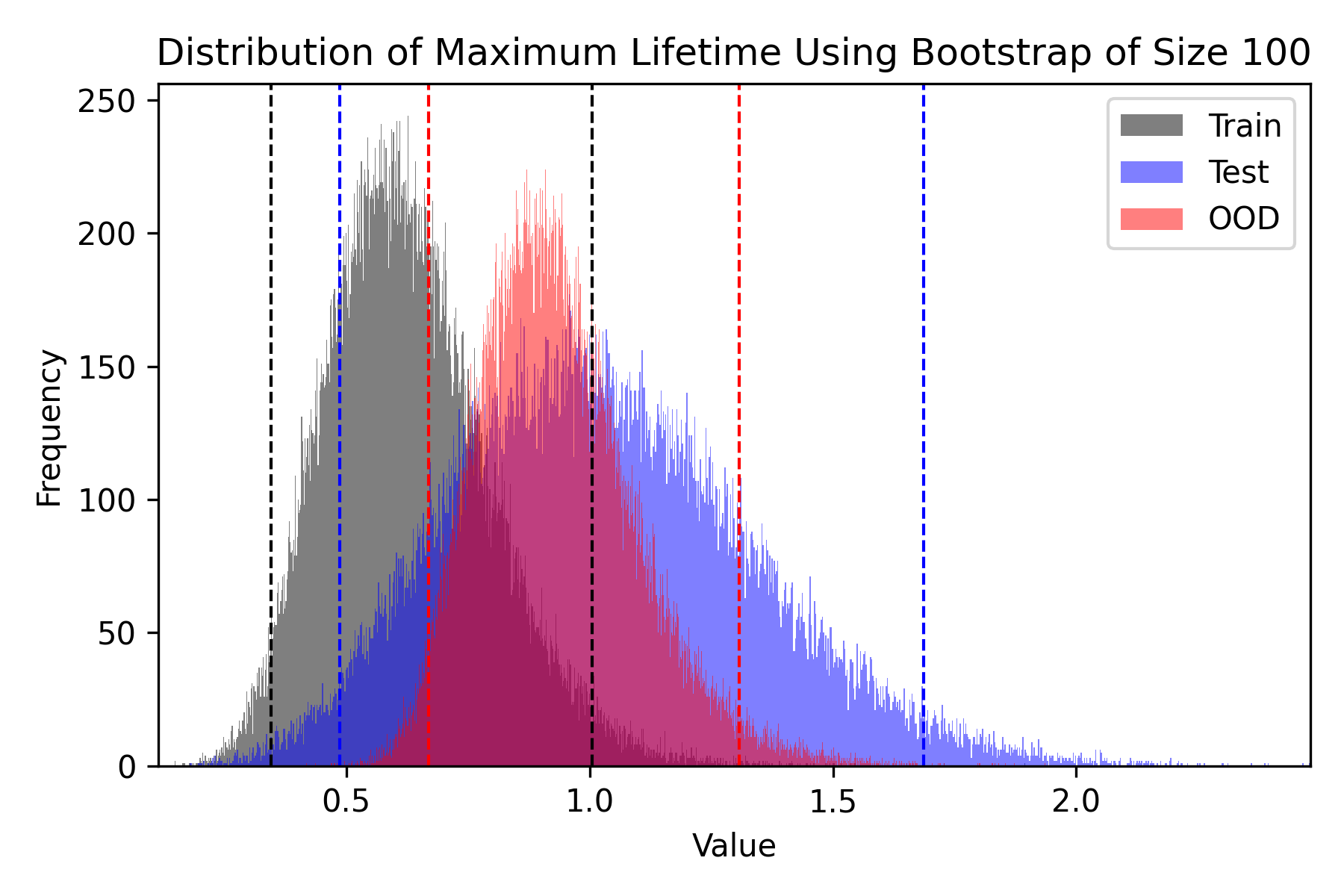}
    \end{subfigure}
    \vspace{-0.5cm}
    \caption{Distribution of average (left) and maximum (right) lifetime of holes ($H_1$) in bootstrapped samples of size 100 from the training and testing CIFAR-10 embeddings and CIFAR-100 embeddings. Vertical lines indicate the 95\% confidence interval.}
\end{figure}

\end{document}